%% file: PaperForReview.tex
\documentclass[10pt,twocolumn,letterpaper]{article}

\usepackage[pagenumbers]{cvpr} 

\usepackage{graphicx}
\usepackage{amsmath}
\usepackage{amssymb}
\usepackage{booktabs}
\usepackage{mathtools}
\usepackage{amsthm}
\usepackage{balance}
\usepackage{microtype}
\usepackage{graphicx}
\usepackage{xcolor}
\usepackage[pagebackref,breaklinks,colorlinks]{hyperref}

\usepackage[capitalize]{cleveref}
\crefname{section}{Sec.}{Secs.}
\Crefname{section}{Section}{Sections}
\Crefname{table}{Table}{Tables}
\crefname{table}{Tab.}{Tabs.}
\input{misc/math_commands}
\input{misc/macro}

\def\cvprPaperID{****} 
\def\confName{CVPR}
\def\confYear{2023}

\begin{document}

\title{Instance-Aware Image Completion}
\author{Jinoh Cho$^1$, Minguk Kang$^1$, Vibhav Vineet$^2$, Jaesik Park$^1$ \\
$^1$ Pohang University of Science and Technology (POSTECH), South Korea\\
$^2$ Microsoft Research, United States\\}
\maketitle

\begin{abstract}
\input{sections/abstract}
\end{abstract}

\input{sections/introduction}
\input{sections/related_work}
\input{sections/method}
\input{sections/experiments}

\input{sections/conclusion}
\clearpage
{\small
\bibliographystyle{ieee_fullname}
\bibliography{egbib}
}
\input{sections/appendix}

\end{document}

%% file: misc/math_commands.tex
\usepackage{amsmath,amsfonts,bm}

\def\eqref#1{equation~\ref{#1}}

\def\1{\bm{1}}

\def\vb{{\bm{b}}}
\def\vc{{\bm{c}}}

\def\vh{{\bm{h}}}

\def\vt{{\bm{t}}}

\def\vz{{\bm{z}}}

\DeclareMathAlphabet{\mathsfit}{\encodingdefault}{\sfdefault}{m}{sl}
\SetMathAlphabet{\mathsfit}{bold}{\encodingdefault}{\sfdefault}{bx}{n}



%% file: misc/macro.tex
\crefname{section}{Sec.}{Secs.}
\Crefname{section}{Section}{Sections}
\Crefname{table}{Table}{Tables}
\crefname{table}{Tab.}{Tabs.}

\newcommand{\Etal}   {et al.~}

\newcommand{\OursAcronym}{\textit{ImComplete}}

\definecolor{yellowone}{rgb}{1.0, 0.88, 0.21}
\definecolor{yellowtwo}{rgb}{0.98, 0.91, 0.71}

\usepackage{times}
\usepackage{epsfig}
\usepackage{graphicx}
\usepackage{amsmath}
\usepackage{amssymb}
\usepackage{verbatim}
\usepackage{hhline}
\usepackage{multirow}
\usepackage{makecell}

\usepackage{ amssymb }
\makeatletter
\newcommand{\ostar}{\mathbin{\mathpalette\make@circled\star}}
\newcommand{\make@circled}[2]{%
  \ooalign{$\m@th#1\smallbigcirc{#1}$\cr\hidewidth$\m@th#1#2$\hidewidth\cr}%
}
\newcommand{\smallbigcirc}[1]{%
  \vcenter{\hbox{\scalebox{0.77778}{$\m@th#1\bigcirc$}}}%
}
\makeatother

\makeatletter
\newcommand\footnoteref[1]{\protected@xdef\@thefnmark{\ref{#1}}\@footnotemark}
\makeatother

\usepackage{xcolor,colortbl}

\usepackage{xspace}

\usepackage{amssymb}
\usepackage{pifont}

\makeatletter
\DeclareRobustCommand\onedot{\futurelet\@let@token\@onedot}
\def\@onedot{\ifx\@let@token.\else.\null\fi\xspace}



%% file: sections/abstract.tex
Image completion is a task that aims to fill in the missing region of a masked image with plausible contents.
However, existing image completion methods tend to fill in the missing region with the surrounding texture instead of hallucinating a visual instance that is suitable in accordance with the context of the scene.
In this work, we propose a novel image completion model, dubbed \OursAcronym{}, that hallucinates the missing instance that harmonizes well with - and thus preserves - the original context.
\OursAcronym{} first adopts a transformer architecture that considers the visible instances and the location of the missing region.
Then, \OursAcronym{} completes the semantic segmentation masks within the missing region, providing pixel-level semantic and structural guidance. 
Finally, the image synthesis blocks generate photo-realistic content. We perform a comprehensive evaluation of the results in terms of visual quality (LPIPS and FID) and contextual preservation scores (CLIPscore and object detection accuracy) with COCO-panoptic and Visual Genome datasets. Experimental results show the superiority of \OursAcronym{} on various natural images.

%% file: sections/introduction.tex
\section{Introduction}
\label{sec:introduction}
\input{assets/figures/teaser}

Image completion is the task of restoring arbitrary missing regions in an image. Researchers have been working on developing image completion models for various applications, such as image editing~\cite{Jo_2019_ICCV, Ling2021EditGANHS}, restoration~\cite{wan2020bringing, liang2021swinir}, and object removal~\cite{shetty2018adversarial}. Even though current image completion models can produce highly realistic results, most previous works focus on filling in missing regions in a realistic way without considering the appropriate instance that needs to be inserted and how it harmonizes with the undamaged regions. For example, we observe that even the cutting-edge image completion models~\cite{yu2018generative, li2022mat, Lugmayr_2022_CVPR} tend to use textures from the surrounding areas to fill in the missing parts rather than synthesizing plausible instances.

When the image completion model fills in a missing region with surrounding textures, it can significantly alter the overall context of the image. For example, the removal of the horse in the image of ~\cref{fig:teaser} changes the context around the missing region from ``a person riding a horse on the beach" to ``a boy walking on the beach". Unfortunately, it is rarely studied to develop an image completion framework that can synthesize the missing region with an instance in a photo-realistic as well as context-conserved fashion.

In this paper, we propose an image completion pipeline, named \OursAcronym{}, that can reason the type of missing instance in the damaged image and complete the damaged image with plausible contents. \OursAcronym{} completes images in three stages: 1) identifying the class of the missing instance, 2) generating a semantic segmentation mask for the missing area, and 3) using the segmentation guidance to complete the missing region. Specifically, \OursAcronym{} employs a transformer network to examine the image's context, defined by analyzing the co-occurrence of instances, to predict the class of the missing instance. Then, it utilizes a conditional GAN and transformer body reconstruction network to generate separate segmentation masks for the missing instance and background of the missing region. Lastly, with state-of-the-art semantic image synthesis approaches~\cite{park2019SPADE, schonfeld2021you, rombach2021highresolution}, \OursAcronym{} fills an instance and background scenes that fit the context in the missing region.

To compare our model against existing image completion methods, we suggest using two evaluation metrics: (1) DETR~\cite{carion2020end} Accuracy to check whether to complete the proper instance (2) CLIPScore~\cite{hessel2021clipscore} to measure how much the context of the image changes after completion by focusing both instance and non-instance parts of the missing regions. We also evaluate our method using standard image quality assessment metrics, FID~\cite{heusel2017gans} and LPIPS~\cite{zhang2018unreasonable}. 
The results on COCO-panoptic~\cite{lin2014microsoft} and Visual Genome~\cite{krishna2017visual} datasets show that \OursAcronym{} has a similar level of visual quality compared to the state-of-the-art image completion approaches. However, the DETR Accuracy and CLIPScore indicate that \OursAcronym{} can complete damaged images better than the other methods with suitable instances. 

\vspace{1.5mm}
\noindent  Our contributions are summarized as follows: 
\begin{itemize}
    \item We propose a new image completion pipeline called \OursAcronym{} that completes the missing region of a masked image in a context-preserved manner.
    \item We propose to use DETR Accuracy and CLIPScore to evaluate the instance/context consistency between the original image and the completed image.
    \item \OursAcronym{} produces high-quality completion results. Our results show better contextural scores (DETR Accuracy and CLIPScore) compared with cutting-edge image completion approaches while keeping comparable image quality scores (FID and LPIPS).
    \item We show that \OursAcronym{} can be plugged with state-of-the-art semantic image synthesis models, such as SPADE~\cite{park2019SPADE}, OASIS~\cite{schonfeld2021you}, and Stable Diffusion~\cite{rombach2021highresolution}.
    \item We show that \OursAcronym{} can perform \emph{object removal} as well by skipping the proposed missing instance inference step, which demonstrates the flexibility of the proposed approach.
\end{itemize}

%% file: assets/figures/Teaser.tex
\begin{figure*}[!ht]
    \vspace{-3mm}
    \includegraphics[width=1\linewidth]{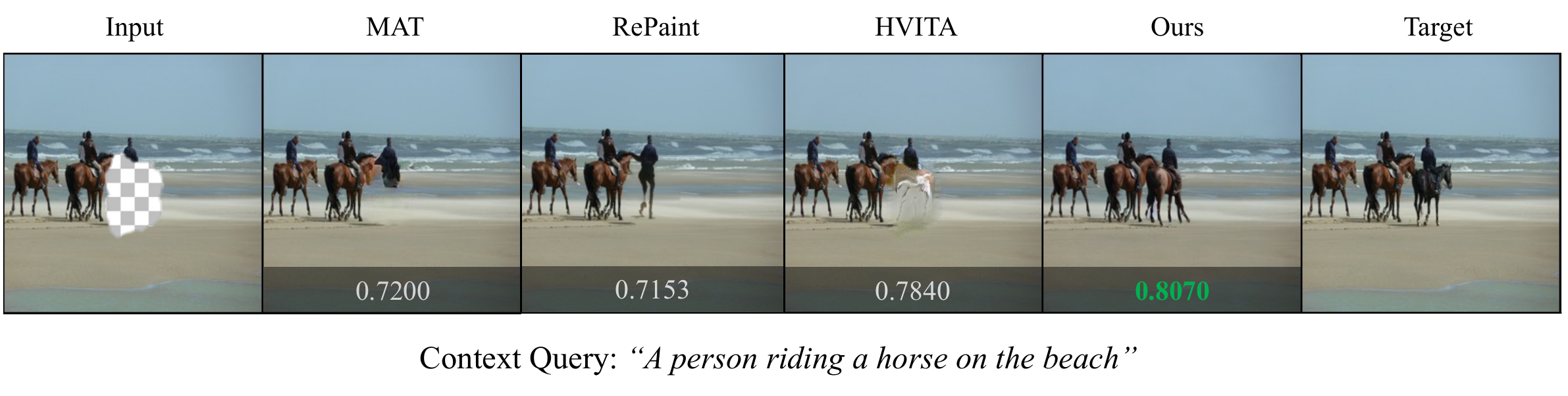}
    \vspace{-3mm}
    \caption{From the first column, Input image with a missing region, results of state-of-the-art image completion approaches, such as MAT~\cite{li2022mat}, RePaint~\cite{Lugmayr_2022_CVPR} and HVITA~\cite{qiu2020hallucinating}, our result~(\OursAcronym{}), and the target image. We compute CLIPScore around the generated part using the query text. As our approach generates a horse to complete the image rather than fills using background textures, CLIPScore of our result exhibits the best performance among the other models.}
    \label{fig:teaser}
\end{figure*}

%% file: sections/related_work.tex
\section{Related Work}
\subsection{Image Completion}
Early research on image completion can be roughly divided into diffusion-based~\cite{bertalmio2000image, ballester2001filling} and patch-based methods~\cite{criminisi2003object, criminisi2004region, ding2018image, hays2007scene, le2011examplar, lee2016laplacian, sun2005image}. These models assume that the missing region of an image can be found and replaced with patterns and features from the remaining parts of the image, making the model fill in the missing region using only basic visual elements and repeating patterns.

With the advance of deep learning, image completion models based on deep generative models have become the mainstream to achieve photo-realistic image completion. Context encoder~\cite{Pathak2016ContextEF} utilizes adversarial training inspired by Generative Adversarial Network (GAN)~\cite{Goodfellow2014GenerativeAN} and shows perceptually faithful completion results. VQ-GAN~\cite{esser2021taming} and MaskGIT~\cite{chang2022maskgit} use auto-regressive generative transformer to fill the missing region. RePaint~\cite{Lugmayr_2022_CVPR} proposes the new inference scheme using pretrained denosing denoising diffusion model~\cite{song2021scorebased, NEURIPS2020_DDPM}, which reduces the distortion between the generated and known region.

Along with introducing advanced generative models, researchers have also been working to improve image completion performance by modifying existing architectures and convolution operations. Series of studies~\cite{song2018contextual,yan2018shift,yu2018generative,liu2019coherent} propose contextual attention layers to encode long-range contextual embeddings and perform image completion based on these. Liu~\Etal~\cite{liu2018image} propose a partial convolutional layer to mitigate color discrepancy and blurriness in completed images. Yu~\Etal~\cite{Yu2019FreeFormII} generalize the partial convolution by introducing a dynamic feature selection mechanism at each spatial coordinate. 
 
 Recent works have focused on completing larger missing region. Zhao~\Etal~\cite{zhao2021comodgan} deal with the challenging image completion by bridging image conditioning with the modulation technique used in StyleGAN2~\cite{karras2019style}. Li~\Etal~\cite{li2022mat} propose a transformer block that can effectively capture long-range context interactions and hallucinate large missing regions.

\subsection{Semantic Image Synthesis}
Generating photo-realistic images from a semantic segmentation mask is called semantic image synthesis. The objective is to generate realistic images that accurately reflect the semantic guideline given by a segmentation mask. To achieve this, the authors of SPADE~\cite{park2019SPADE} develop a spatially-adaptive (de)normalization layer to modulate semantic information to the pixel-wise image features. Besides, OASIS~\cite{schonfeld2021you} improves the power of the discriminator by replacing the vanilla discriminator with a segmentation-based discriminator. This replacement allows the generator to be trained with more informative signal from the discriminator, resulting in better synthesis results. SPADE and OASIS were initially not designed to perform the image completion task. However, in this paper, we extend the usage of semantic image synthesis blocks and use them for semantic-guided image completion where the segmentation guidance is created by our proposed instance segmentation generator and background segmentation completion network (depicted in Fig.~\ref{fig:overallarchitecture}).

Moreover, Latent Diffusion Model (LDM) and Stable Diffusion~\cite{rombach2021highresolution} utilize powerful cross-attention mechanism~\cite{vaswani2017attention} to condition the diffusion model on various inputs, such as texts and segmentation masks. To leverage the large-scale, pretrained stable diffusion model for our task, we devise a new inference scheme to generate target instances in a missing region using a segmentation mask provided by our framework. Similar to our approach, SPG-Net~\cite{song2018spg} and SG-Net~\cite{liao2020guidance} use semantic segmentation labels to fuel more informative supervision for image completion. Our approach is similar to SPG-Net and SG-Net in that \OursAcronym{} predicts/recovers the semantic segmentation mask and generate the content based on the mask, but differs in that \OursAcronym{} handles more challenging scenarios where the instances are \emph{entirely removed}.

To our best knowledge, the only existing work for instance-aware image completion is HVITA~\cite{qiu2020hallucinating}, where a target instance is wholly removed from an image. HVITA consists of four steps: (1) detecting visible instances, (2) constructing a graph using detected instances to understand the scene context, (3) generating a missing instance and placing it on the missing region, and (4) refining the inserted image. 
Yet, HVITA depends on conventional object detection and is designed for completing rectangular regions. Despite the additional refinement module in HVITA, it still produces low-quality image completion results. In particular, distortion occurs at the boundary between the generated instance and its surroundings.
On the other hand, our proposed \OursAcronym{} is free to handle arbitrarily shaped masks. Our sophisticated pipeline for segmentation mask recovery helps to understand the context of images better and encourages visual continuity and plausibility on the boundaries between generated and unmasked regions. 

%% file: sections/method.tex
\section{Method}
\label{sec:method}

Our framework aims to complete the corrupted image, where the visual instance is completely removed. 
This problem is challenging as the model must not only generate a target instance but also ensure that the generated instance seamlessly blends with the remaining areas of the image.
To alleviate the problem, our framework~(\OursAcronym{}) completes masked images in three steps. 
First, we infer a contextually appropriate instance by figuring out the category of the missing instance (\cref{sec:method:missing_instance_prediction}). 
Second, we complete the semantic segmentation map of the missing region based on the inference result from the previous step (\cref{sec:method:segmentation_map_generation}). 
Finally, we transform the masked image with semantic segmentation maps to a realistic completed image (\cref{sec:method:image_completion}). 

\input{assets/figures/OverallArchitecture.tex}

The overview figure of our framework is shown in \cref{fig:overallarchitecture}. We leave out the all the architecture details and training details in the \cref{appendix:architecture and training details}.

\subsection{Missing Instance Prediction}
\label{sec:method:missing_instance_prediction}

To obtain the relationship information between instances in a given scene, \OursAcronym{} first predicts instance bounding box coordinates, instance classes, and a semantic segmentation map using the pre-trained DETR~\cite{carion2020end}. Let the panoptic segmentation map $\boldsymbol{S}_{\text{M}} = \text{DETR}(\boldsymbol{I}_{M})$, then we can extract box coordinates of the visible instances $\boldsymbol{B} = [\vb_{1}, ..., \vb_{k}]$ and object classes $\vc = [c_{1}, ..., c_{k}]^{\top}$ from $\boldsymbol{S}_{\text{M}}$, where $k$ is the number of predicted instances. Then, to infer the class of the missing instance $y_{target}$,  a transformer network, called missing instance inference transformer utilizes tokens obtained from the object classes $\vc$ of visible instances, as well as additional tokens responsible for the missing region. In particular, we convert the visible instances' classes  $\vc$  into learnable input tokens using a single linear layer. A quick approach is to utilize object queries from DETR as input tokens directly, but we observe such a method exhibits a worse performance compared to employing new learnable class embeddings. 

Furthermore, to inject the location information of the visible instances, we embed their bounding box coordinates into positional encoding vectors and sum them to the learnable class embeddings. To acquire the positional encoding vectors, we input the normalized center coordinate~($C_x$, $C_y$), width~($W$), and height~($H$) of the bounding box to a single linear layer with sigmoid activation function. We also apply the same procedure for the missing region token that will be used for missing instance inference. We explored different ways to create the positional encoding vectors in \cref{appendix:pe_variants} and adopt the aforementioned positional encoding since it gives the best missing instance inference performance. The below formulations are mathematical expressions of how our missing instance infer transformer works.

\begin{equation}
    \begin{aligned}
        \vz_{0} & = \textbf{E}_{class} + \textbf{E}_{pos}\\
                & = \text{MLP(}\vc^\prime) + \sigma(\text{MLP(}\boldsymbol{B}^\prime)), & \vz_{0} \in \mathbb{R}^{(k + 1) \times d}\\
        \vz_{l}^{\prime} & =
        \text{MSA}(\text{LN}(\vz_{l-1})) + \vz_{l-1}, & \quad l = 1, ..., L
        \\
        \vz_{l} & = \text{MLP}(\text{LN}(\vz_{l}^{\prime})) + \vz_{l}^{\prime}, & l = 1, ..., L
        \\
        y & = \text{LN}(\vz^{0}_{L}), & 
    \end{aligned}
\end{equation}


\noindent  where MLP is a multi-layer perceptron, $\sigma$ is a sigmoid activation, $d$ is the dimension of embedding vectors, $\boldsymbol{B}^\prime$ is $[\vb_{0}] \cup \boldsymbol{B}$, $\vb_{0}$ is the missing region bounding box coordinate, and $\vc^\prime$ is $[c_{0}] \cup \vc$ where $c_{0}$ is an extra class token for missing instance.
The missing instance inference transformer consists of 12 transformer encoder layers ($L = 12$) with eight heads. The missing region token interacts with the visible region tokens through self-attention mechanism. Thus, the network can more accurately predict the likely class of the missing instance based on the detected instances and their location information. Additional information regarding the training and architecture specifics can be found within \cref{appendix:mint architecture}.

\subsection{Semantic Segmentation Map Generation}
\label{sec:method:segmentation_map_generation}
Utilizing the predicted class of the missing instance from the previous step, we aim to generate the semantic segmentation map of the missing region. We create the segmentation map of the instance and the background area individually with separate modules (instance segmentation generator and background segmentation completion network) and obtain the final segmentation map by inserting the missing instance segmentation into the background segmentation, as shown in \cref{fig:overallarchitecture}.

\paragraph{Instance Segmentation Generator.}
We perform the missing instance segmentation generation using two modules: a generator and a discriminator, for instance segmentation. The instance segmentation generator aims to create a plausible segmentation map corresponding to the predicted instance class. For the implementation, we use the architecture from BigGAN~\cite{Brock2019LargeSG}, one of the most successful conditional GANs, with slight modification. We input the predicted missing instance class from the previous step and the box coordinates of the missing region to the Conditional Batch Normalization~\cite{de2017modulating} module in the instance segmentation generator. We train the model by using spectral normalization~\cite{Miyato2018SpectralNF} and hinge loss~\cite{Lim2017GeometricG} with DiffAug~\cite{zhao2020diffaugment} technique. 

\paragraph{Background Segmentation Completion Network.}
The background segmentation completion network produces the segmentation map of the non-instance region without attempting to generate the missing instance. To do this, we randomly scribble the ground truth segmentation map and let the background segmentation completion network restore it using cross-entropy loss. We experimentally identify that this procedure successfully reconstructs the background segmentation maps. The background segmentation completion network is implemented using convolutional heads and tails with a transformer body. In \cref{sec:experiment:bsc_comparsion}, we demonstrate the importance of transformer body, particularly in cases where there exists a large hole in the damaged image. 

Finally, we obtain the overall segmentation map of the missing region by inserting the instance segmentation into the background segmentation. Further training and architecture details can be found in \cref{appendix:isg architecture} and \cref{appendix:bsc architecture}.

\subsection{Segmentation-guided Image Completion}
\label{sec:method:image_completion}

This module is designed to complete the masked image using the reconstructed segmentation mask as the guidance. There are three versions~(\OursAcronym$_{spade}$, \OursAcronym$_{oasis}$, and \OursAcronym$_{stable}$) depending on the image generation approach plugged into our framework. 

\paragraph{SPADE/OASIS Version.}
For the \OursAcronym$_{spade}$ and \OursAcronym$_{oasis}$, 
we can use UNet~\cite{ronneberger2015u}-like completion model to hallucinate the missing region using the masked image and predicted segmentation map pairs. The input to the completion model is the masked image, and the predicted segmentation map is conditioned by SPADE~\cite{park2019SPADE} or OASIS~\cite{schonfeld2021you} blocks to help the image completion model precisely fill in the missing region based on the semantics that segmentation map provides. We only apply the conditioning blocks to the decoder part of the completion network. For \OursAcronym$_{oasis}$, the generator and discriminator losses proposed in the OASIS paper are applied to the masked region, and L2 loss is penalized to the remaining undamaged area to synthesize a plausible instance while preserving contents of the undamaged region. Moreover, to learn specific modulation only on the mask regions, we assign an extra class label to the masked regions of the segmentation maps. We leave out the architecture and training details in \cref{appendix:sgc architecture}.

\paragraph{Stable Diffusion Version.}
\input{assets/figures/OverviewStable}
In the case of \OursAcronym$_{stable}$, we use the pretrained text-conditioned stable diffusion inpainting model~\cite{rombach2021highresolution} to incorporate powerful image generation ability into the instance-aware image completion task. We experimentally identify that the instance-aware completion task using stable diffusion can be accomplished by conducting consecutive image completion processes on each segment class from the restored segmentation map.

Let's denote the entire segmentation mask in the missing region as $\boldsymbol{S_{miss}} = \{(S_{1}, c_{1}), ..., (S_{n}, c_{n})\}$, where n is the number of segment classes predicted, and $c_{n}$ is the class of each segment $S_{n}$. Then, we can formulate the segmentation-guided image completion mechanism of \OursAcronym$_{stable}$ as follow:
\begin{equation}
    \begin{aligned}
        I_{n} & = G(I_{n-1}, S_{n}, T(c_{n})), & n = 1, ..., N
    \end{aligned}
\end{equation}
where \textit{G} is the pretrained stable diffusion inpainting model, $I_{0}$ is an original damaged image, and \textit{T} is a function that converts the class label to the corresponding text prompt. The instance-aware completion approach of \OursAcronym$_{stable}$ is illustrated in \cref{fig:stable_diff_method}.

Since \OursAcronym{} can complete a masked image according to the arbitrary segmentation map, \OursAcronym{} can reconstruct diverse target instances of the same category by simply feeding different segmentation maps into the segmentation-guided image completion model. \Cref{fig:multiple_solution} visually demonstrates that \OursAcronym{} can synthesize instances of various shapes within the same class.

%% file: assets/figures/OverallArchitecture.tex
\begin{figure*}[!t]
    \vspace{-3mm}
    \includegraphics[width=1\linewidth]{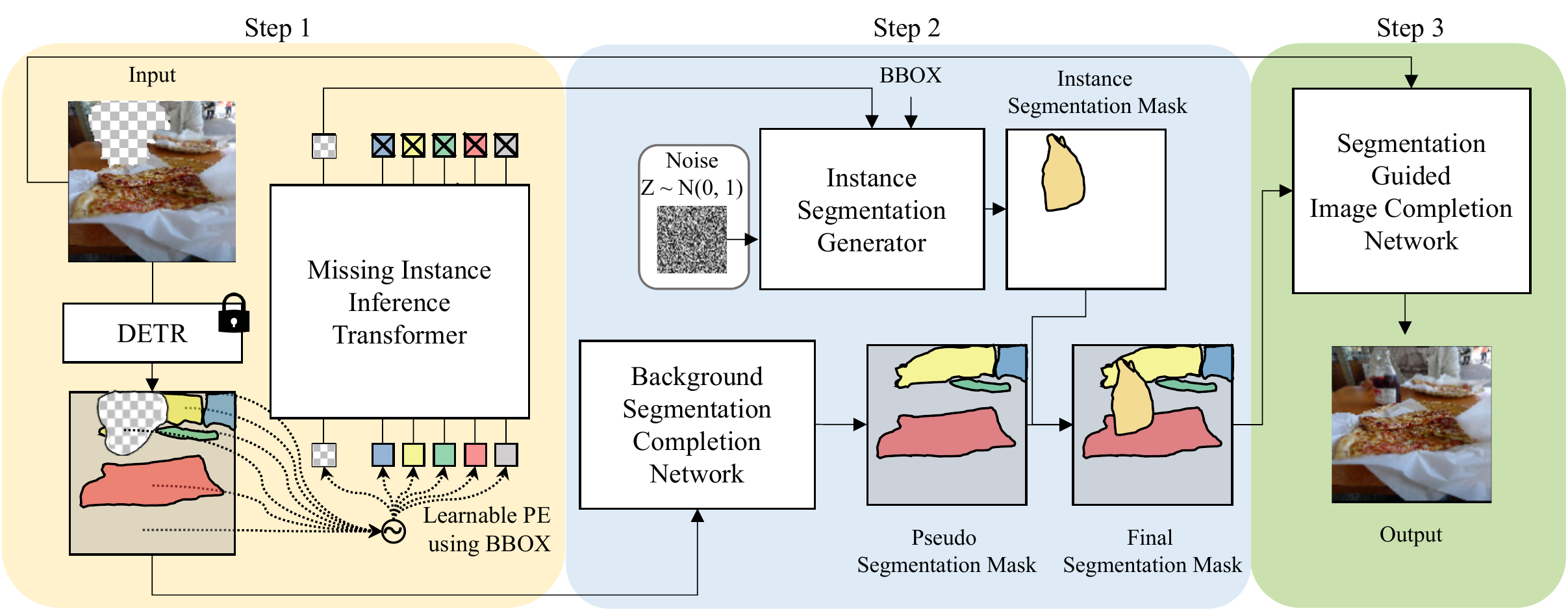}
    \caption{Overview of the proposed approach, called \OursAcronym{}. \OursAcronym{} completes the image in three steps: (1) infer the missing instance class, (2) complete a segmentation map in the missing region, and (3) translate the segmentation map to an image to hallucinate the missing region.
    }
    \vspace{-3mm}
    \label{fig:overallarchitecture}
\end{figure*}

%% file: assets/figures/OverviewStable.tex
\begin{figure}[!ht]
    \vspace{-5mm}
    \includegraphics[width=1\linewidth]{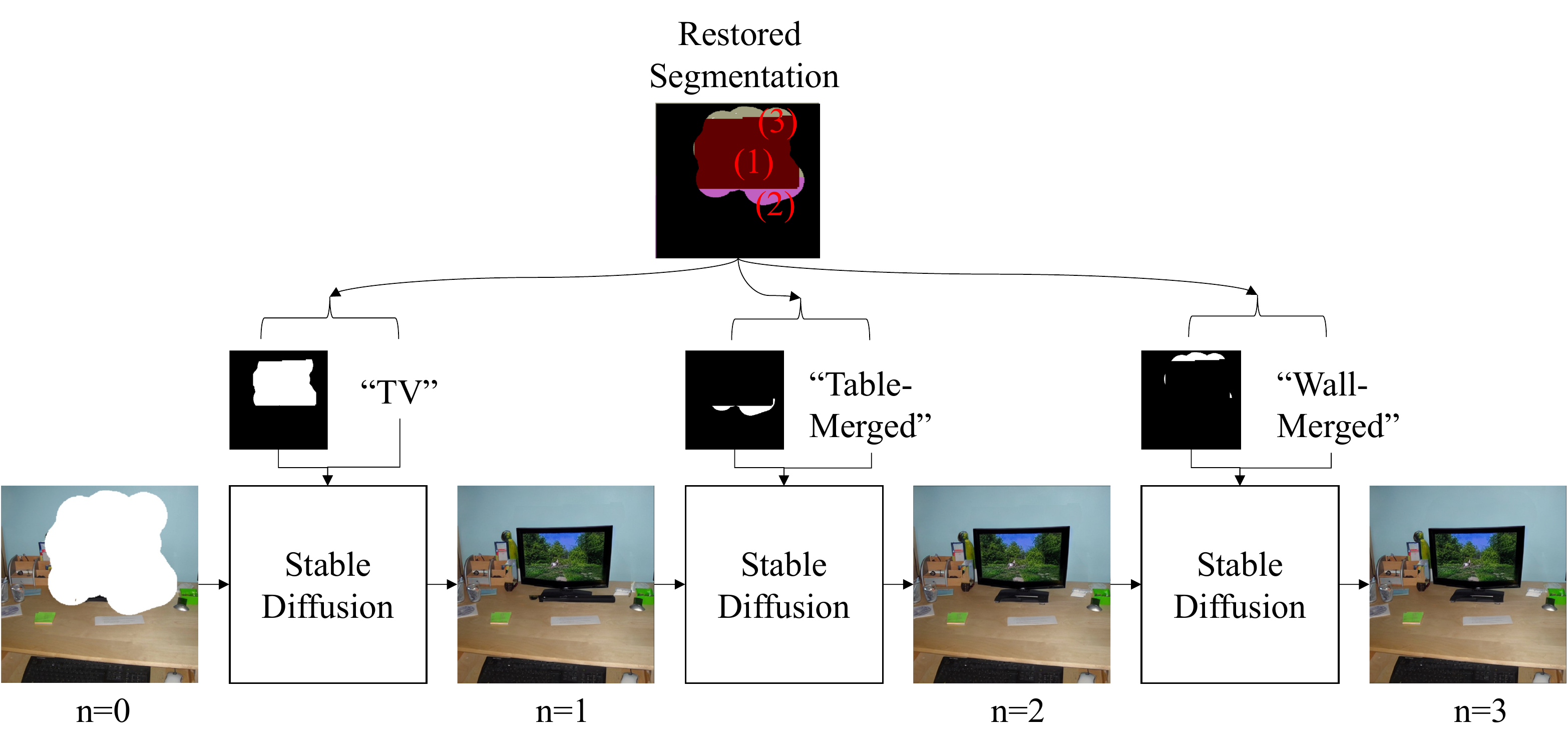}
    \caption{\OursAcronym{} modifies the standard stable diffusion inpainting inference scheme. In each step, the diffusion model iteratively inpaints the masked region for the given text prompt and segment mask.}
    \vspace{-3mm}
    \label{fig:stable_diff_method}
\end{figure}

%% file: sections/experiments.tex
\section{Experiments}
\subsection{Datasets}
\label{sec:experiment:datasets} 
By following HVITA~\cite{qiu2020hallucinating}, a closest work to ours, we use COCO-panoptic~\cite{lin2014microsoft} and Visual Genome~\cite{lin2014microsoft} dataset for the demonstration.

\vspace{1.5mm}
\noindent  \textbf{COCO-panoptic.} The dataset~\cite{lin2014microsoft}  contains 118K images for the train set and 5K images for the validation set with 80 things classes and 91 stuff classes. Compared with center-aligned datasets e.g., FFHQ~\cite{karras2019style}, CelebA-HQ~\cite{liu2015deep}, and ImageNet~\cite{deng2009image}, COCO-panoptic is more challenging, multiple instances appear, and hardly applied for image completion. For creating training data, we make a missing region using the following procedures. (0) We re-split original train and validation set to increase the number of validation set. (1) We select a rectangular region that contains an instance. (2) We randomly crop the image to be from 10\% to 50\% of the whole image, including the missing rectangular region. (3) We randomly pick 50 points around the bounding box and draw thick lines between those points to make irregular regions. We select 30 classes from the things classes of COCO-panoptic that are frequently observed in the images.

\vspace{1.5mm}
\noindent  \textbf{Visual Genome.}
We adopt Visual Genome~\cite{krishna2017visual} dataset to check the generalization ability of image completion methods. The dataset contains 110k images with fine-grained 34k object categories. Due to the category mismatch between COCO-panoptic and Visual Genome, we use DETR-predicted results as the pseudo annotations. Then, we generate missing regions using the detected object boxes with the same approach for creating the COCO-panoptic dataset.

\subsection{Baselines}
\label{sec:experiment:baselines} 

\noindent \textbf{HVITA}~\cite{qiu2020hallucinating} is closely related work considering our problem setup. We could not find any publicly available code to reproduce results in the paper, so we implemented HVITA from scratch by carefully referring to the details in the paper~\cite{qiu2020hallucinating}.

\vspace{1.5mm}
\noindent \textbf{MAT}~\cite{li2022mat} is one of the state-of-the-art image completion models built on GAN framework. It consists of newly designed transformer blocks with a style manipulation module and mask updating strategy. We use the official author's implementation for training and evaluation.

\vspace{1.5mm}
\noindent \textbf{RePaint}~\cite{Lugmayr_2022_CVPR} is a cutting-edge diffusion-based image completion model. It introduces a novel inference algorithm to condition the given masked image, which is suitable for the image completion task by utilizing pre-trained unconditional DDPM~\cite{NEURIPS2020_DDPM}. We use the author's official implementation for training and evaluation.
\input{assets/figures/MainImageResult.tex}
\input{assets/tables/MainNumberTable.tex}

\subsection{Metrics}
\label{sec:experiment:metrics} 
\noindent \textbf{LPIPS and FID.} There are several metrics for image quality measurement. L1, MSE, and PSNR quantify a pixel-wise error, and SSIM calculates the patch-wise similarity between a generated image and the original one using luminance, contrast, and structure information. These metrics, however, are known to be inconsistent with human perception~\cite{zhang2018unreasonable}. In addition, since there are various possible answers to complete the missing region, L1, MSE, and PSNR may show inappropriate values, although completion results are plausible. To overcome this issue, we use LPIPS~\cite{zhang2018unreasonable}, which is well known to agree with human perception by computing L2 distance on learned feature space~(ImageNet-trained VGG~\cite{simonyan2014very}), not on the pixel space. Moreover, we utilize Fr\'echet Inception Distance~(FID)~\cite{heusel2017gans} to assess the realism of completed images. 

\vspace{1.5mm}
\noindent \textbf{DETR Acc.}
\label{sec:experiment:metrics:DETR}
To quantify whether the generated instances preserve the original context of damaged images, we use pre-trained DETR to detect instances for the given images and check the equality between generated instances' class and the target instances' class. 

\vspace{1.5mm}
\noindent \textbf{CLIPScore.}
\label{sec:experiment:metrics:CLIPScore}
The generated instances are expected to be of high quality and blend seamlessly with the unmasked area. To achieve this, both the instance and non-instance parts of the missing region should be synthesized faithfully. Therefore, it is necessary to evaluate image completion models by considering both the instance and non-instance parts. However, the DETR Acc metric tends to focus on the target instance class, as shown in \cref{appendix:need_clipscore}. In order to achieve realism and harmonization between masked and unmasked areas, we suggest using CLIPScore~\cite{hessel2021clipscore}, which helps evaluate how well the hallucinated area is perceptually aligned with the given text description.

CLIPscore has recently been shown to quantify text-to-image alignment and visual quality together~\cite{saharia2022photorealistic, yu2022scaling}. We employ a pretrained CLIP ViT-B/32 visual and textual encoders~\cite{radford2021learning} and compute CLIPScore by calculating the cosine similarity between the embeddings of the completed region and text description. Since the COCO-panoptic and Visual Genome datasets do not provide captions for the images, we obtain captions by feeding the original real images to the high-performance captioning models, OFA~\cite{wang2022unifying}. CLIP score can be formulated as follows:

\begin{equation}
    \begin{aligned}
       \operatorname{CLIPScore}(\vh, \vt) = 2.5\operatorname{max}(\operatorname{\text{cos}}(\vh, \vt), 0),
    \end{aligned}
\end{equation}
where $\text{cos}(\cdot,\cdot)$ means cosine similarity operation, $\vh$ indicates the image embedding of hallucinated region, and $\vt$ represents CLIP text embedding.

\subsection{Evaluation Results}

To show the strengths of \OursAcronym{}, we perform the image completion task on the masked COCO-panoptic dataset (\cref{sec:experiment:datasets}). We train/finetuned $\text{MAT}_{pre}$~\cite{li2022mat}, $\text{MAT}_{scratch}$~\cite{li2022mat}, RePaint~\cite{Lugmayr_2022_CVPR}, HVITA~\cite{qiu2020hallucinating}, and our model using the dataset except the segmentation guided image completion network of \OursAcronym$_{stable}$. We evaluate the models using a test split of the masked COCO-panoptic dataset. 

The results are summarized in \cref{tbl:main_result}. RePaint shows the best LPIPS and FID scores of 0.084 and 6.511, respectively. But it exhibits inferior results in CLIPscore and DETR Acc in which \OursAcronym{}$_{stable}$ gives the best results (CLIPscore is 0.663, and VGA is 40.221\%). HVITA can generate the target instance better than MAT and RePaint. Still, the generated instances are less realistic than the other models, as seen in  \cref{tbl:main_result} and \cref{fig:main_image_result}. \OursAcronym{}$_{stable}$ is the model that compensates for the shortcomings of HVITA but reinforces the strengths. \OursAcronym{}$_{stable}$ attains comparable FID of 6.714 to the FID of RePaint~(6.511).

To test the generalization ability of each model, we evaluate the COCO-trained completion models on Visual Genome in the zero-shot fashion. As can be seen in \cref{tbl:main_result}, \OursAcronym{}$_{stable}$~ proves the excellence in CLIPScore and DETR Acc metrics. Note that high LPIPS does not necessarily indicate low-fidelity results because the results may include instances that differ from those of the original images as shown in \cref{fig:multiple_solution} and \cref{fig:failure_result}. Since LPIPS performs pixel-wise evaluation in the feature space, the model must generate an inpainting similar to the original image to obtain a low LPIPS score. However, our model can complete the missing regions with visual instances of various shapes and classes, likely to differ from the original instance. Such diverse outputs result in comparatively higher LPIPS scores.

\input{assets/figures/MultipleSolution.tex}
\input{assets/figures/BSCResult.tex}
\subsection{Background Completion}

\label{sec:experiment:bsc_comparsion}
Inspired by MAT~\cite{li2022mat}, which uses a transformer architecture in the body of the network to complete large-scale masked images, we adopt the transformer architecture for the body of the U-Net like background segmentation completion network. As shown in \cref{fig:bsc_combined}, the transformer architecture helps reconstruct background semantic labels, especially for the large hole in the segmentation. As the hole size in segmentation from DETR increases, the performance gap in mIoU between transformer- and conv-body background segmentation completion networks becomes more stand out. We also find that the conv-body network generates artifacts in the missing background segmentation in large hole settings, as shown in \cref{fig:bsc_combined}. This demonstrates that the transformer architecture is better suited for recovering missing segmentation by leveraging global-range context interaction. Further architecture and training details can be found in \cref{appendix:bsc architecture}.

\subsection{Object Removal}
\label{sec:experiment:removal}
\input{assets/figures/ObjRemovedImage.tex}

\input{assets/tables/ObjRemovalNumberTable.tex}

While our work primarily focuses on generating a plausible instance in the missing region, object removal - which focuses on overwriting the region with natural background - is also widely studied in the field of image completion. In this regard, we conducted experiments to verify if our model can perform object removal. From the main framework presented in \cref{fig:overallarchitecture}, we can skip the pipeline for predicting the missing instance in step 1 and use only the background segmentation map from step 2 to complete the missing region.

As shown in \cref{fig:obj_removal_image}, the qualitative results demonstrate the capability of \OursAcronym{} to remove unwanted instances without significant degradation in visual quality. Moreover, \Cref{tbl:obj_removal_fid} presents FID values for our vanilla model and the modified model designed for object removal. In comparison to the model for instance completion, the object removal version of \OursAcronym{} exhibits a similar level of visual quality.

\subsection{Failure Cases}
\label{sec:experiment::failure}
\input{assets/figures/FailureResult.tex}

\OursAcronym{} may fail to generate correct class instances due to the wrong predicted class from the missing instance inference transformer. However, as shown in Figure~\ref{fig:failure_result}, the generated instances are well harmonized with the remaining parts. From the first and second rows, \OursAcronym{} generates a cake and a cat instead of a dog. The images are still convincing.

%% file: assets/figures/MainImageResult.tex
\begin{figure*}[!t]
\vspace{-3mm}
\includegraphics[width=1\linewidth]{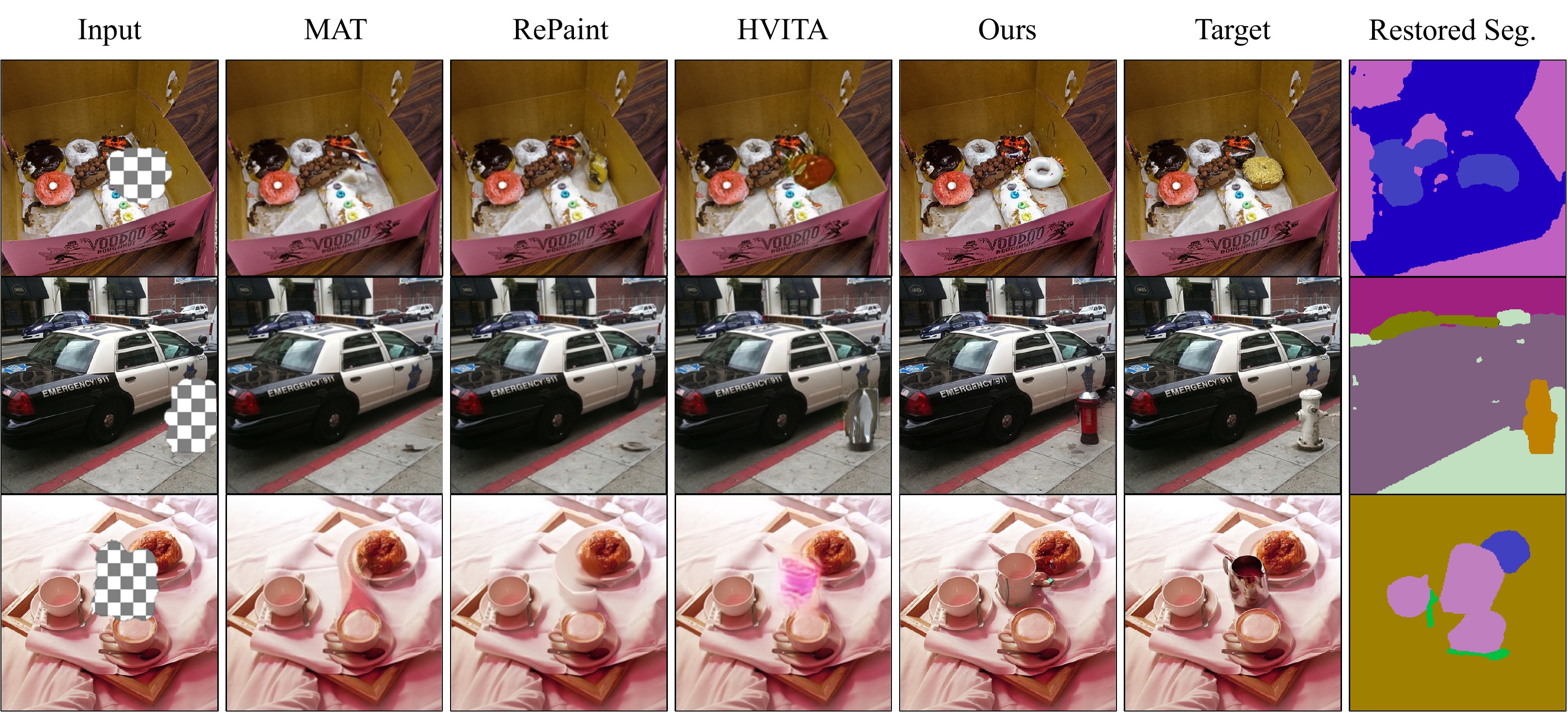}
\vspace{-6mm}
\caption{From the left side: Input, MAT~\cite{li2022mat}, RePaint~\cite{Lugmayr_2022_CVPR}, HVITA~\cite{qiu2020hallucinating}, Ours, Target, Restored Seg from \OursAcronym. We leave out more results on the Figure~\ref{fig:more_coco_1} and~\ref{fig:more_coco_2} in \cref{appendix:additional_result}.}
\label{fig:main_image_result}
\end{figure*}

%% file: assets/tables/MainNumberTable.tex
\begin{table*}[!h]
\caption{Comparison of image synthesis quality on COCO-panoptic and Visual Genome datasets. All models are trained/finetuned on COCO-panoptic and evaluated on COCO-panoptic and Visual Genome (Zero-shot) images. We compare the quality of generated images using 4 metrics: CLIPScore, DETR Acc, LPIPS and FID. MAT$_{pre}$/RePaint~\cite{Lugmayr_2022_CVPR} is pretrained on Places365-Standard~\cite{zhou2017places} that contains 8M images and is finetuned with our modified COCO dataset. MAT$_{scratch}$ and the other models are trained from scratch only using the modified COCO. We mark the best, the second-best in \colorbox{yellowone}{normal yellow} and \colorbox{yellowtwo}{light yellow} respectively.}
\setlength\tabcolsep{4.0pt}
\label{tbl:main_result}
\resizebox{1\textwidth}{!}{
    \setlength{\arrayrulewidth}{0.9pt}
    \begin{tabular}{l|cccc|cccc}
    & \multicolumn{4}{c|}{} 
    & \multicolumn{4}{c}{}
    \\
    [-0.8em]
    \multirow{2}*[-0.5ex]{\large{Metric}}
    & \multicolumn{4}{c|}{COCO-panoptic} 
    & \multicolumn{4}{c}{Visual Genome (Zero-shot)}
    \\
    & \multicolumn{4}{c|}{} 
    & \multicolumn{4}{c}{}
    \\
    [-0.8em]
    \cline{2-9}
    & \multicolumn{4}{c|}{} 
    & \multicolumn{4}{c}{}
    \\
    [-0.8em]
    & CLIPscore~$\uparrow$ 
    & DETR Acc~$(\%)$~$\uparrow$
    & LPIPS~$\downarrow$
    & FID~$\downarrow$
    & CLIPscore~$\uparrow$
    & DETR Acc~$(\%)$~$\uparrow$
    & LPIPS~$\downarrow$
    & FID~$\downarrow$
    \\ [2pt]
    \hline
    & \multicolumn{4}{c|}{} 
    & \multicolumn{4}{c}{}
    \\
    [-0.8em]
    MAT$_{pre}$~{\small\cite{li2022mat}}
    &0.614
    &1.894
    &\cellcolor{yellowtwo}0.087
    &7.192 & 0.615 & 1.815 &\cellcolor{yellowtwo}0.087 & 6.488
    \\
    
    MAT$_{scratch}$~{\small\cite{li2022mat}}
    &0.606
    &1.910
    &0.093
    &7.895 & 0.608 & 1.766 & 0.091 & 7.302
    \\

    RePaint~{\small\cite{Lugmayr_2022_CVPR}}
    &0.635
    &9.350
    &\cellcolor{yellowone}0.084
    &\cellcolor{yellowone}6.511 &0.638 & 7.273 & \cellcolor{yellowone}0.083 & \cellcolor{yellowone}5.123
    \\
    \hline

    HVITA~{\small\cite{qiu2020hallucinating}}
    &0.567
    &16.626
    &0.132
    &10.496 & 0.568 & 18.900 & 0.129 & 9.311
    \\
    \hline
    \OursAcronym$_{spade}$
    &0.626
    &33.551
    &0.122
    &8.519 & 0.628 & 35.123 & 0.120 & 6.891
    \\
    
    \OursAcronym$_{oasis}$
    &\cellcolor{yellowtwo}0.641
    &\cellcolor{yellowtwo}36.832
    &0.119
    &8.284 & \cellcolor{yellowtwo}0.644 & \cellcolor{yellowtwo}38.224 & 0.116 & 6.549
    \\

    \OursAcronym$_{stable}$
    &\cellcolor{yellowone}0.663
    & \cellcolor{yellowone} 40.221
    &0.089
    &\cellcolor{yellowtwo}6.714 & \cellcolor{yellowone}0.659 &  \cellcolor{yellowone}41.014 & 0.089 & \cellcolor{yellowtwo}6.221
    \\
    \hline
    \end{tabular}
}

\end{table*}

%% file: assets/figures/MultipleSolution.tex
\begin{figure}[!h]
    \vspace{-3mm}
    \includegraphics[width=1\linewidth]{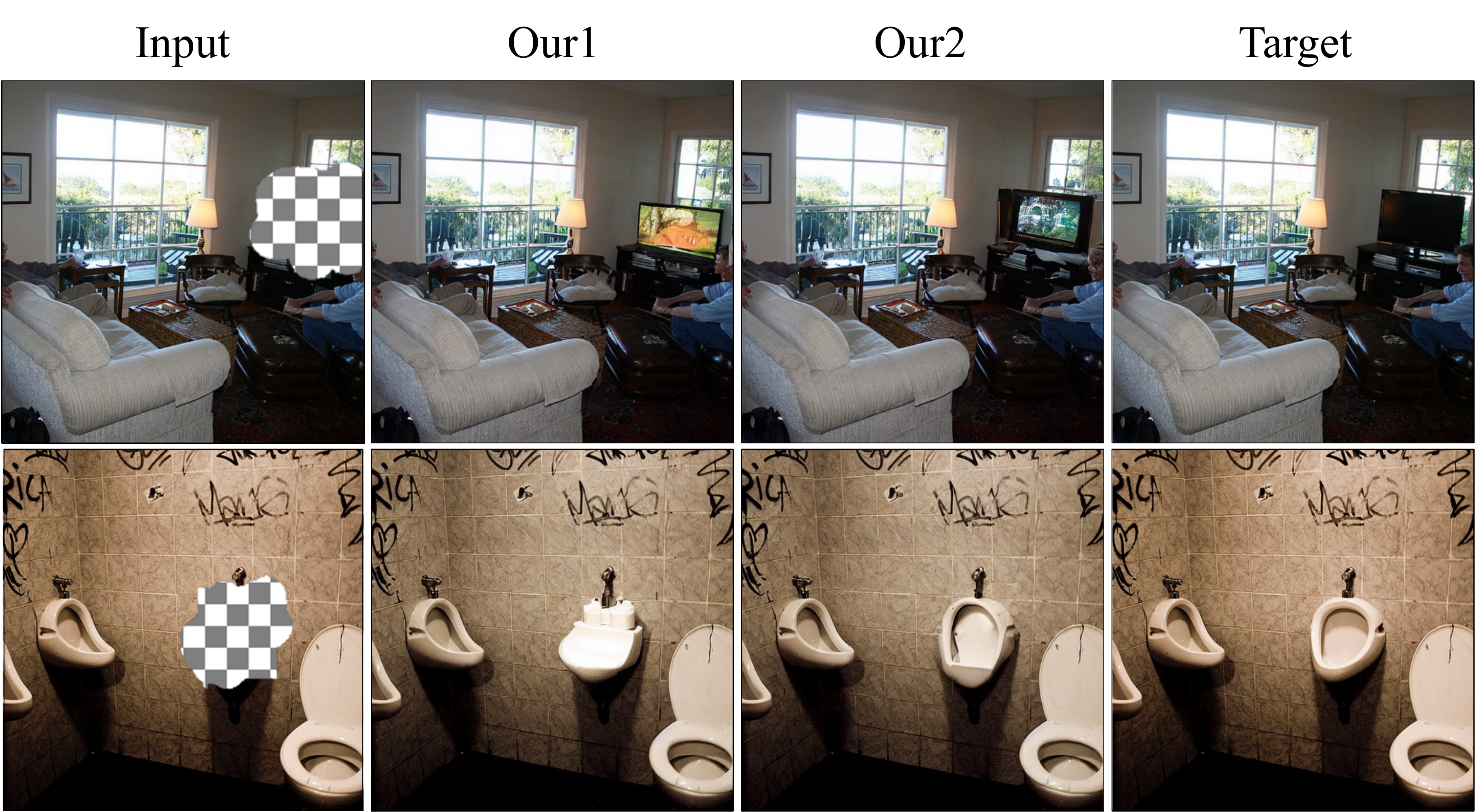}
    \caption{Example of diverse completion results of the proposed approach.}
    \vspace{-3mm}
    \label{fig:multiple_solution}
    \vspace{-0.3cm}
\end{figure}

%% file: assets/figures/BSCResult.tex
\begin{figure*}[!t]
    \vspace{-10mm}
    \includegraphics[width=1\linewidth]{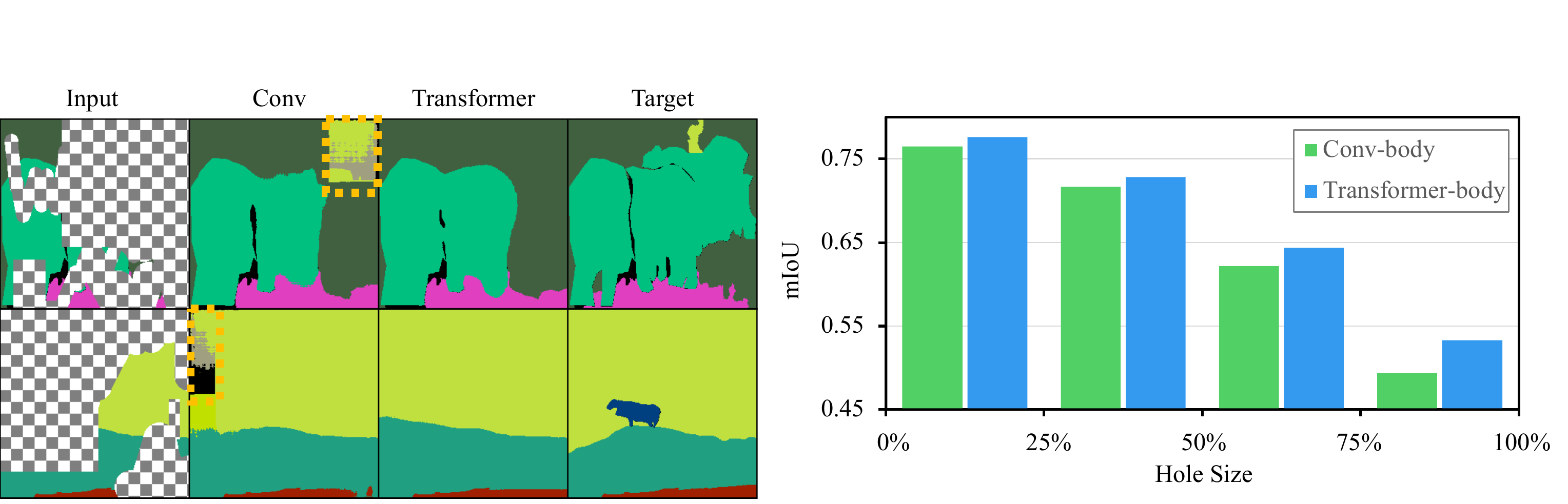}
    \vspace{-6mm}
    \caption{Predicting semantic masks given the severe missing regions. Transformer-body background completion network is more robust to the hole size than the convolution-body version that often shows unwanted artifacts, as highlighted in yellow dotted boxes. The hole sizes of the first and the second row examples are 76.4$\%$ and 71.3$\%$, respectively.}
    \label{fig:bsc_combined}
\end{figure*}

%% file: assets/figures/ObjRemovedImage.tex
\begin{figure}[!h]
    \vspace{-6mm}
    \includegraphics[width=1\linewidth]{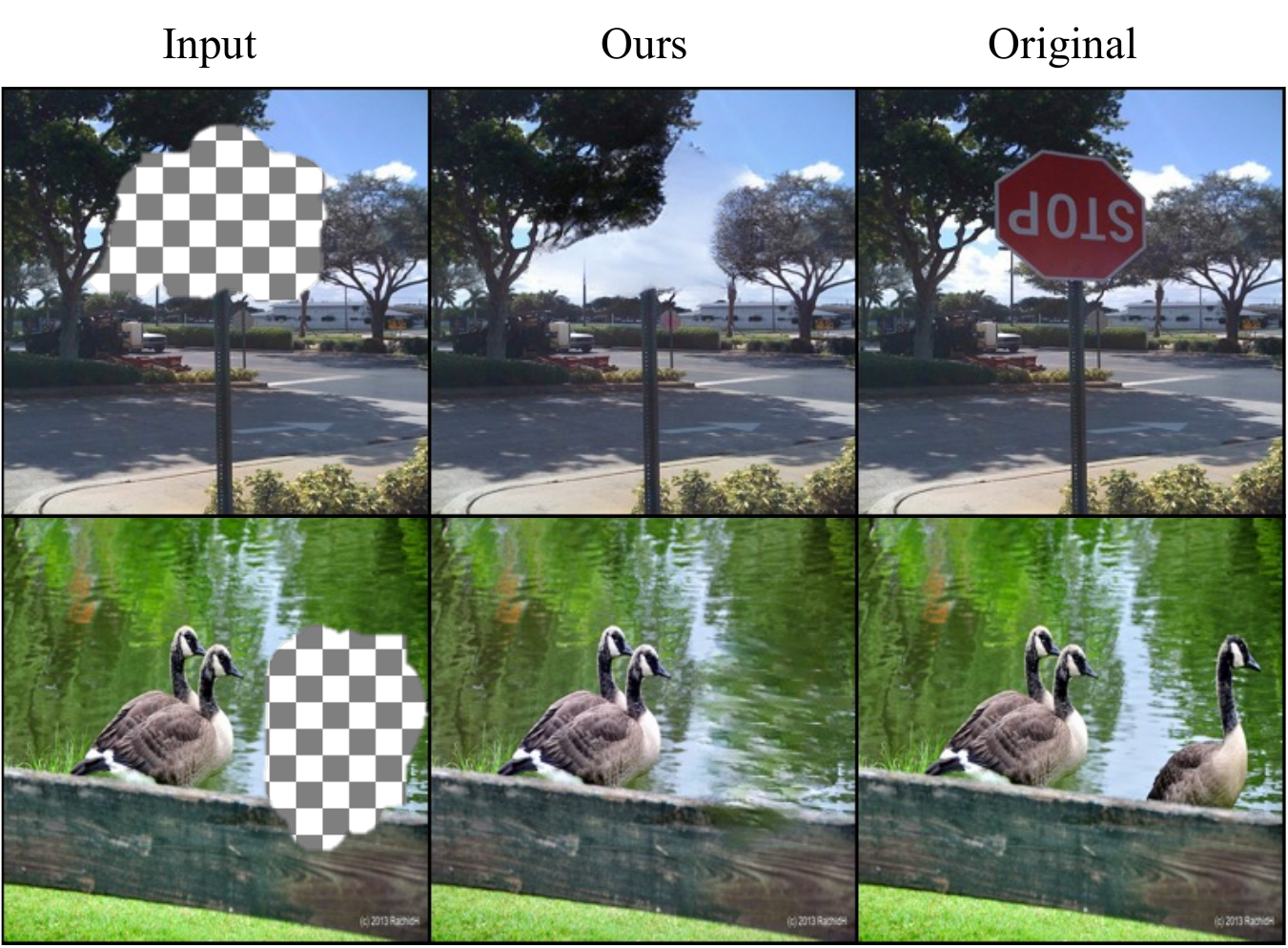}
    \vspace{-6mm}
    \caption{Results of \OursAcronym{}'s ``\emph{no instance}" version. \OursAcronym{} can perform \emph{object removing} by simply skipping the foreground object prediction step in the proposed pipeline. }
    \vspace{-3mm}
    \label{fig:obj_removal_image}
\end{figure}

%% file: assets/tables/ObjRemovalNumberTable.tex
\begin{table}[!h]
\caption{\label{tbl:obj_removal_fid} FID comparison between vanilla model in \OursAcronym{} and ``\emph{no instance}" version proposed in Sec~\ref{sec:experiment:removal}.}
\resizebox{1.0\linewidth}{!}{
    \begin{tabular}{c|cc}
        & vanilla  & ``\emph{no instance}"\\
        \hline
         COCO-panoptic & 7.284 & 7.874~(\textcolor{purple}{$\uparrow$~0.590}) \\
         Visual Genome~(Zero-shot) & 6.849 & 7.887~(\textcolor{purple}{$\uparrow$~1.038}) \\
        \hline
    \end{tabular}
}
\end{table}

%% file: assets/figures/FailureResult.tex
\begin{figure}[ht]
    \centering
    \vspace{-6mm}
    \includegraphics[width=1\linewidth]{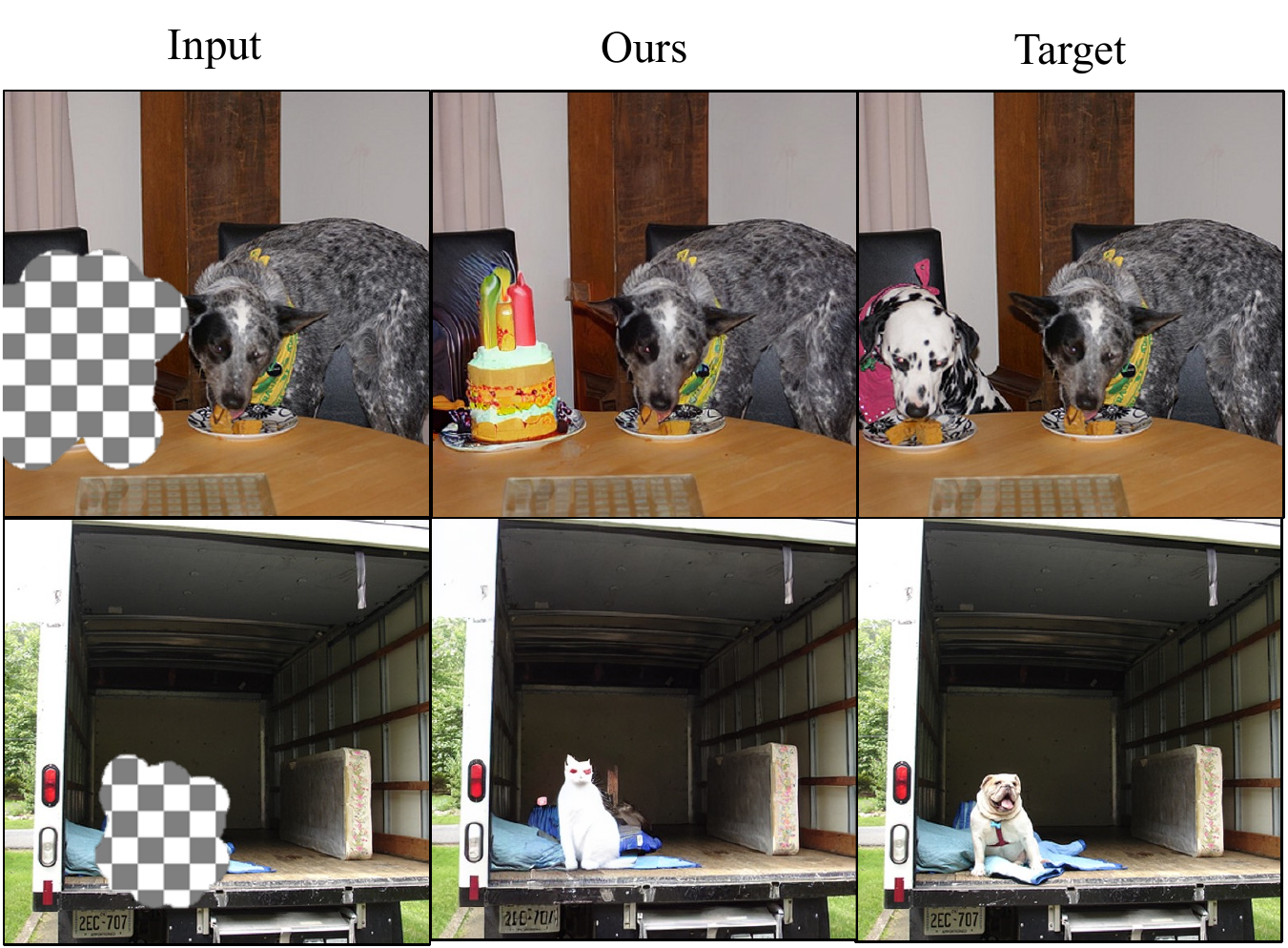}
    \vspace{-6mm}
    \caption{\label{fig:failure_result}Failure cases of \OursAcronym{}. \OursAcronym{} approach may fail to predict the same class of the target image. However, the generated images still follow the scene context.}
    \vspace{-3mm}
\end{figure}

%% file: sections/conclusion.tex
\section{Conclusion}
\label{sec:conclusion}
This paper presents a novel framework called \OursAcronym{} for instance-aware image completion. \OursAcronym{} can synthesize context-friendly visual instances instead of filling in the missing region with surrounding textures. Through a number of experiments, we have demonstrated that \OursAcronym{} has an advantage over the baseline models in terms of CLIPScore and DETR Acc, while still being comparable to the baseline models in terms of FID scores. Moreover, \OursAcronym{} can perform both object removal and completion with the proper instances. \OursAcronym{} also can easily integrate cutting-edge segmentation-guided image completion model.

\vspace{1.5mm}
\noindent \textbf{Limitation.}
Our approach is demonstrated with the 30 classes of COCO-panoptic things classes. Rather than using pre-defined categories, open-set segmentation and text-to-image translation would be an interesting direction. 

\vspace{1.5mm}
\noindent \textbf{Acknowledgement.} This work was supported by Institute of Information \& communications Technology Planning \& Evaluation (IITP) grant funded by the Korea government(MSIT) (No.2019-0-01906, Artificial Intelligence Graduate School Program(POSTECH) and No.2021-0-00537, a technology that restores invisible parts in the video with visual common sense through self-directed learning).

%% file: sections/appendix.tex
\appendix
\onecolumn

\section{Using CLIPScore for Image Completion Evaluation}
\label{appendix:need_clipscore}
We aim to complete images in the missing region by generating context-friendly instances. However, if the background region of the missing region is not properly restored, the visual discontinuity will occur, leading to unnatural images. Using the DETR prediction result, we may judge whether the generated instances are convincing. However, as the DETR only focuses on the instances, DETR prediction cannot be solely used to determine whether the background region of the missing region has been restored well. To handle the issue, we propose to use CLIPScore. While DETR only perceives the class label of the masked region, such as ``dog", CLIPScore can consider both the instance class and its surroundings by using the detailed description~(Context Query) of the masked region, such as ``a black and white dog walking in the dirt." (See the Figure~\ref{fig:clip_need_method} and~\ref{fig:clip_need}.)

To experimentally determine if the CLIPScore can evaluate the generated instance and its backgrounds in the missing region, we first complete images with target segmentation. We swap the background regions of the completed images with random background images using the non-instance region masks. Next, we extract the context queries characterizing the masked regions from the original images using a pre-trained image captioning model~\cite{wang2022unifying}. Finally, we calculate the CLIPScore and verify the DETR detection result. We only add up the CLIPScore of generated/swapped images where DETR's prediction matches the target instance class. As shown in Table~\ref{tbl:clip_need}, the CLIPScore of completed images are higher than the swapped images.

\input{assets/figures/WhyNeedClipMethod.tex}
\input{assets/figures/WhyNeedClipImage.tex}
\input{assets/tables/WhyNeedClipTable.tex}

\newpage
\section{Positional Encoding for Missing Instance Prediction}
\label{appendix:pe_variants}
To predict the class of instances in the missing region, we tested six positional encoding variants for the proposed Missing Instance Inference Transformer. Let's denote relative bounding box coordinate as follow: $R_x=C_x-M_x$ and $R_y=C_y-M_y$, where $(M_x, M_y)$ and $(C_x, C_y)$ are missing region center coordinate and detected instances' bounding box center coordinate, respectively. $H$ and $W$ indicate the width and height of the bounding box. All the variables mentioned above are normalized to [0, 1]. ABS4C represents using $C_x, C_y, H, W$. ABS2C only uses $C_x$ and $C_y$. REL4C represents using $R_x, R_y, H$ and $W$. REL2C only uses $R_x, R_y$. No PE represents not using any positional encoding methods. Learnable means using the learnable positional encoding method. GCN indicates graph classification module in HVITA~\cite{qiu2020hallucinating}. 

Among six methods, using the ABS4C method shows the best performance. Also, our missing instance inference transformer with ABS4C positional encoding shows 4.5$\%$ increase in classification accuracy performance compared to the No PE model. Moreover, using our transformer with ABS4C positional encoding significantly improves the performance of the GCN module in HVITA at 17$\%$. See the details of six variants and their performance comparison in Table~\ref{tbl:mit_number}. 

\input{assets/tables/MITNumberTable.tex}
\section{Analysis of Missing Instance Inference Transformer}
\label{appendix:mit_properties}
In this section, we perform diverse experiments to identify the effectiveness and the generalizability of the proposed missing instance inference module. We first visualize the intermediate self-attention layers to investigate the relations between surrounding instances and the predicted instance. As shown in Figure~\ref{fig:attn_matrix_vis}, when our module predicts a missing class, the module gives more weight to the semantically related surroundings. We then perform another experiment to check whether our module’s prediction also depends on the position of the missing region. Figure~\ref{fig:position_aware_prediction} describes how predicted classes change over different missing regions. From these two observations, we empirically conclude that our module rationally produces missing instance's classes depending on the category and position of surrounding instances.

For the generalizability of our missing instance inference transformer, we adopt a zero-shot setting and see if our module trained on the COCO-panoptic dataset succeeds in producing an accurate prediction class label of the unseen dataset, Visual Genome. Experiment results in Table~\ref{tbl:mit_number} show that the prediction accuracy of the ABS4C method still ranked the best in an unseen dataset. Thus, we conclude our missing instance inference transformer generalizes well to unseen datasets as long as they share the same class labels with the training set.
\input{assets/figures/AttnVisualization.tex}
\input{assets/figures/PositionAwarePrediction.tex}

\clearpage
\section{Additional Results}
\label{appendix:additional_result}
\input{assets/figures/MoreCOCO_1.tex}
\input{assets/figures/MoreCOCO_2.tex}

\section{Network Architectures and Training Details}
\label{appendix:architecture and training details}
We implement our completion model for experiments using PyTorch~\cite{NEURIPS2019_9015} library. We use PyTorch functions and define some necessary notations here to provide the architectural details of \OursAcronym{}.

\textsc{Embedding~($dim_{out}$)} indicates PyTorch embedding function. \textsc{TransEncLayer ($dim_{token}, dim_{hidden}, head_{numb}$)} is a vanilla transformer encoder layer. \textsc{FC~($dim_{input}, dim_{output}$)} is a single linear layer. \textsc{Conv~($kernel_{size}, stride, padding$)} is a convolution layer. \textsc{DBlock~($channel_{in}, channel_{out}, downsampling$)} and \textsc{BigGBlock~($channel_{in}, channel_{out}, upsampling, dim_{z_{split}}, dim_{shared}$)} are BigGAN blocks implemented in StudioGAN libary~\cite{kang2022studiogan}. \textsc{Self-Attention} is a self-attention block implemented in StudioGAN. \textsc{BN} and \textsc{LN} indicate a batch normalization and layer normalization layer, respectively. \textsc{SPADEResnetBlock} is a SPADE block~\cite{park2019SPADE}. \textsc{ResBlock-Up} and -\textsc{Down} indicate ResNet blocks implemented in StudioGAN. \textsc{Up($size$)} indicates a upsampling function PyTorch provides. N is the number of classes in the segmentation mask. m and k indicate the batch size and the number of visible instances for a given masked image.

\subsection{missing instance inference transformer}
\label{appendix:mint architecture}
Missing instance inference transformer consists of 12 layers of transformer encoder layers~\cite{vaswani2017attention} with 8 heads. The architectural details are described in Table~\ref{network:mit}. We use Adam optimizer~\cite{Kingma2015AdamAM} with $\beta_{1}$ and $\beta_{2}$ of 0.9 and 0.999, respectively. The learning rates linearly increase until 50 epochs and linearly decrease until the end of the training~(250 epochs).

\subsection{instance segmentation generator and discriminator}
\label{appendix:isg architecture}
We use BigGAN architecture~\cite{Brock2019LargeSG} implemented in the StudioGAN library
~\cite{kang2022studiogan}. The detailed BigGAN structure can be summarized in Tables~\ref{network:isg_gen} and~\ref{network:isg_dis}, respectively.
We use Adam optimizer with $\beta_{1}$ and $\beta_{2}$ of 0.0 and 0.999, respectively. The learning rates for the generator and discriminator are set to $5.0\times10^{-5}$ and $2.0\times10^{-4}$.

\subsection{background segmentation completion network}
\label{appendix:bsc architecture}
For the convolution-body background segmentation completion network, we use a UNet~\cite{ronneberger2015u} architecture. For the transformer-body background segmentation completion network, we build 8 layers of the Transformer encoder and add 3 convolution layers and 3 ResNet blocks before and after the transformer encoder layers. The details of the transformer-body version are shown in Table~\ref{network:bsc}. We optimize the whole segmentation completion network with Adam optimizer with a learning rate of $1.0\times10^{-4}$ and weight decay of $1.0\times10^{-4}$. $\beta_{1}$ and $\beta_{2}$ for Adam are set to 0.9 and 0.999, respectively. 

\subsection{segmentation-guided completion network}
\label{appendix:sgc architecture}
We use a UNet-like architecture for the segmentation-guided completion network. We apply \textsc{SPADEResnetBlock} to the decoder part to condition the completed segmentation mask into the UNet-like completion network. The architectural details of OASIS~\cite{schonfeld2021you} version of segmentation-guided completion network are explained in Tables~\ref{network:isg_gen} and \ref{network:isg_dis}. 
We follow the same training details as the original OASIS paper used with a single exception of the objective function. In order to preserve the unmasked region's data and generate contents only in the masked region, we apply the OASIS generator and discriminator loss only to the masked region and use additional L2 loss only on the unmasked region. We also add perceptual loss computed on the pre-trained VGG-19 recognition model to encourage fast convergence. For the SPADE~\cite{park2019SPADE} version, we change the OASIS discriminator into the SPADE discriminator. Then, we follow the same training scheme described in the original SPADE paper. 

\input{assets/tables/MITArchitecture.tex}
\input{assets/tables/ISGArchitecture.tex}
\input{assets/tables/BSCArchitecture.tex}
\input{assets/tables/SISDArchitecture.tex}
\input{assets/tables/SISGArchitecture.tex}

%% file: assets/figures/WhyNeedClipMethod.tex
\begin{figure}[h]
    \begin{center}
    \includegraphics[width=1\linewidth]{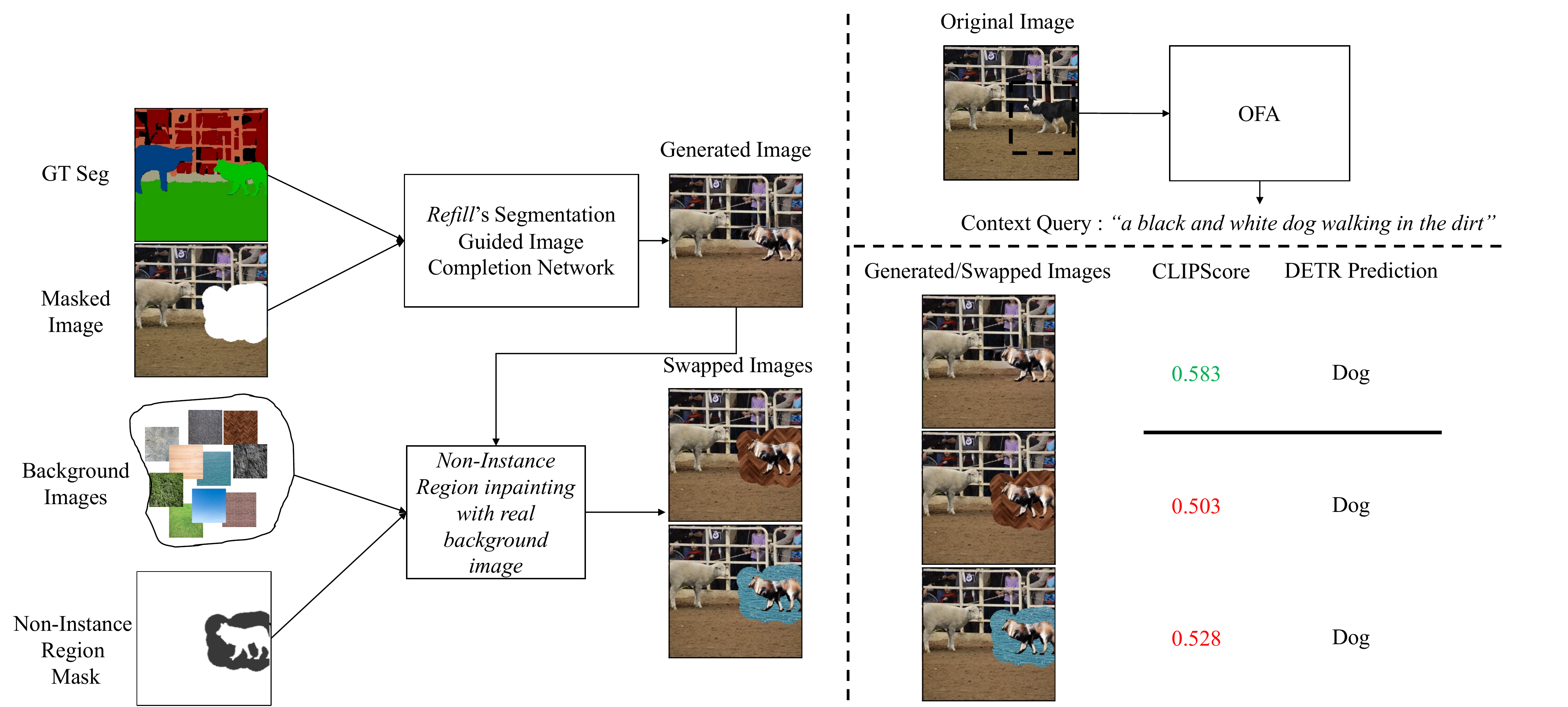}
    \end{center}
    \vspace{-4mm}
    \caption{
     Effectiveness of CLIPScore. We generate images that complete the missing region using the proposed approach. We also create swapped images by replacing backgrounds. Given the context query obtained from the original image, the CLIPScore is highest on the generated images, compared to the CLIPScore of background-swapped images. We can see that properly predicting semantic context can be evaluated via CLIPScore.
    }
    \label{fig:clip_need_method}
\end{figure}

%% file: assets/figures/WhyNeedClipImage.tex
\begin{figure}[h]
    \includegraphics[width=1\linewidth]{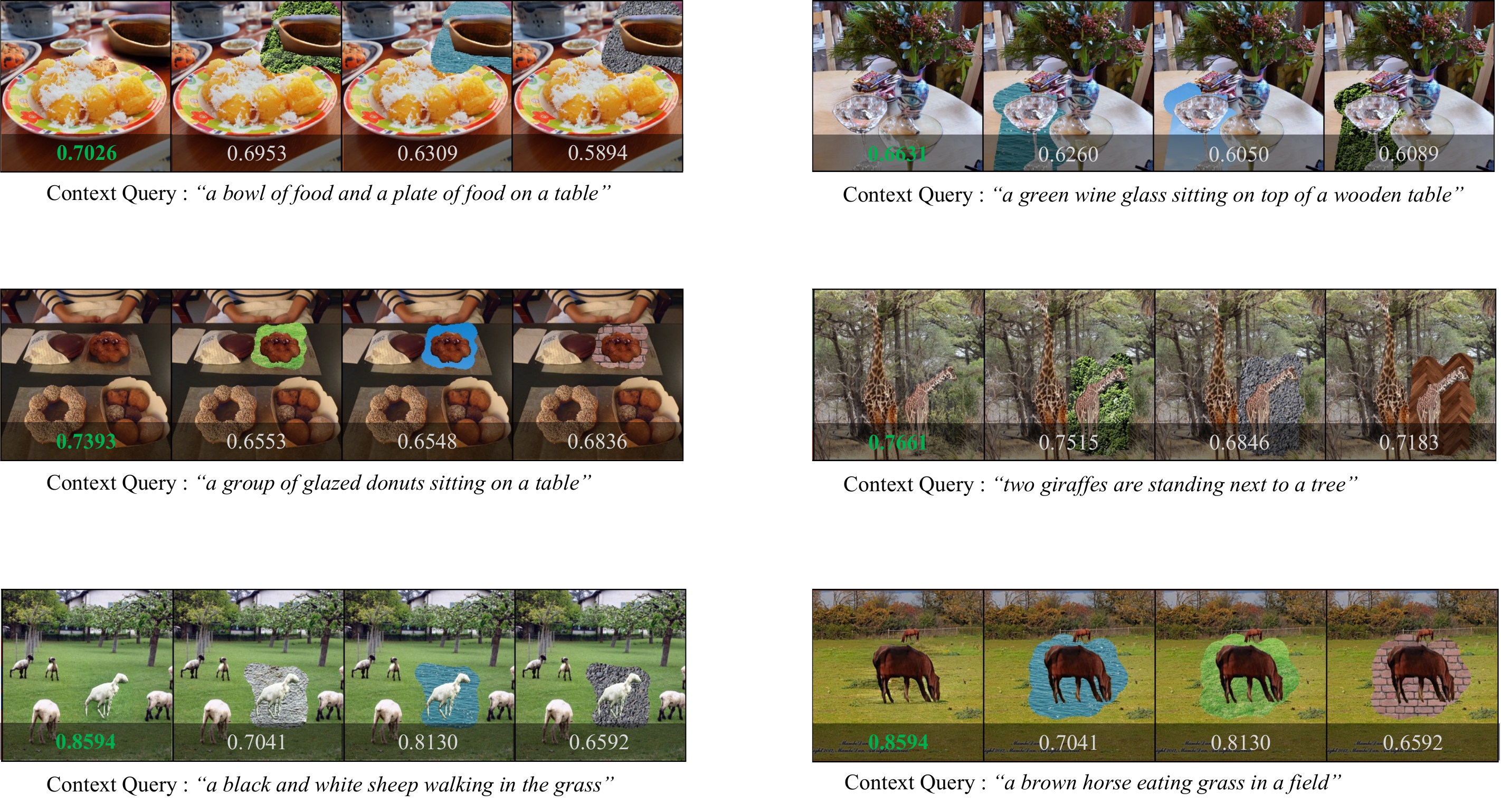}
    \vspace{-3mm}
    \caption{
    For each subfigure, from left to right:  generated images using the predicted segmentation map and background-swapped images. We notice the CLIPScores of the generated images are the highest.}
    \label{fig:clip_need}
\end{figure}

%% file: assets/tables/WhyNeedClipTable.tex
\begin{table}[ht]
\caption{\label{tbl:clip_need} CLIPScore comparison between generated images using our approach and the background-swapped images.}
\centering
    \begin{tabular}{c|cc|cc}
        \noalign{\smallskip}\noalign{\smallskip}\hline
        \multirow{2}{*}{} & \multicolumn{2}{c|}{COCO-panoptic} & \multicolumn{2}{c}{Visual Genome} \\
        \cline{2-5}
              &  Complete  & Swap & Complete & Swap \\
        \hline
         CLIPScore & 0.6855 & 0.6270 & 0.6910 &0.6333 \\
        \hline
    \end{tabular}
\end{table}

%% file: assets/tables/MITNumberTable.tex
\begin{table}[ht]
\caption{\label{tbl:mit_number} Missing instance infer transformer performance on COCO-panoptic and VG~(Zero-shot). The explanation of each methods including ABS4C, REL4C, ABS2C, REL2C, Learnable, No PE and GCN are described in Appendix Sec.~\ref{appendix:pe_variants}.}
\setlength\tabcolsep{4.0pt}
\centering
    \begin{tabular}{l|c|c}
        \toprule
        &
        COCO-panoptic&
        Visual Genome~(Zero-shot)
        \\
        \midrule
        ABS4C
        &67.548
        &67.236~(\textcolor{purple}{$\downarrow$~0.312})
        \\
        
        REL4C
        &66.466
        &66.283~(\textcolor{purple}{$\downarrow$~0.183})
        \\
    
        ABS2C
        &65.204
        &66.055~(\textcolor{teal}{$\uparrow$~1.149})
        \\
        
        REL2C
        &65.865
        &66.370~(\textcolor{teal}{$\uparrow$~0.505})
        \\
        
        Learnable
        &62.981
        &65.701~(\textcolor{teal}{$\uparrow$~2.720})
        \\
    
        No PE
        &63.041
        &65.047~(\textcolor{teal}{$\uparrow$~2.006})
        \\
        
        GCN
        &50.661
        &49.178~(\textcolor{purple}{$\downarrow$~1.483})
        \\      
        \bottomrule
    \end{tabular}

\end{table}

%% file: assets/figures/AttnVisualization.tex
\begin{figure}[!ht]
    \vspace{-2mm}
    \centering
    \includegraphics[width=0.5\textwidth]{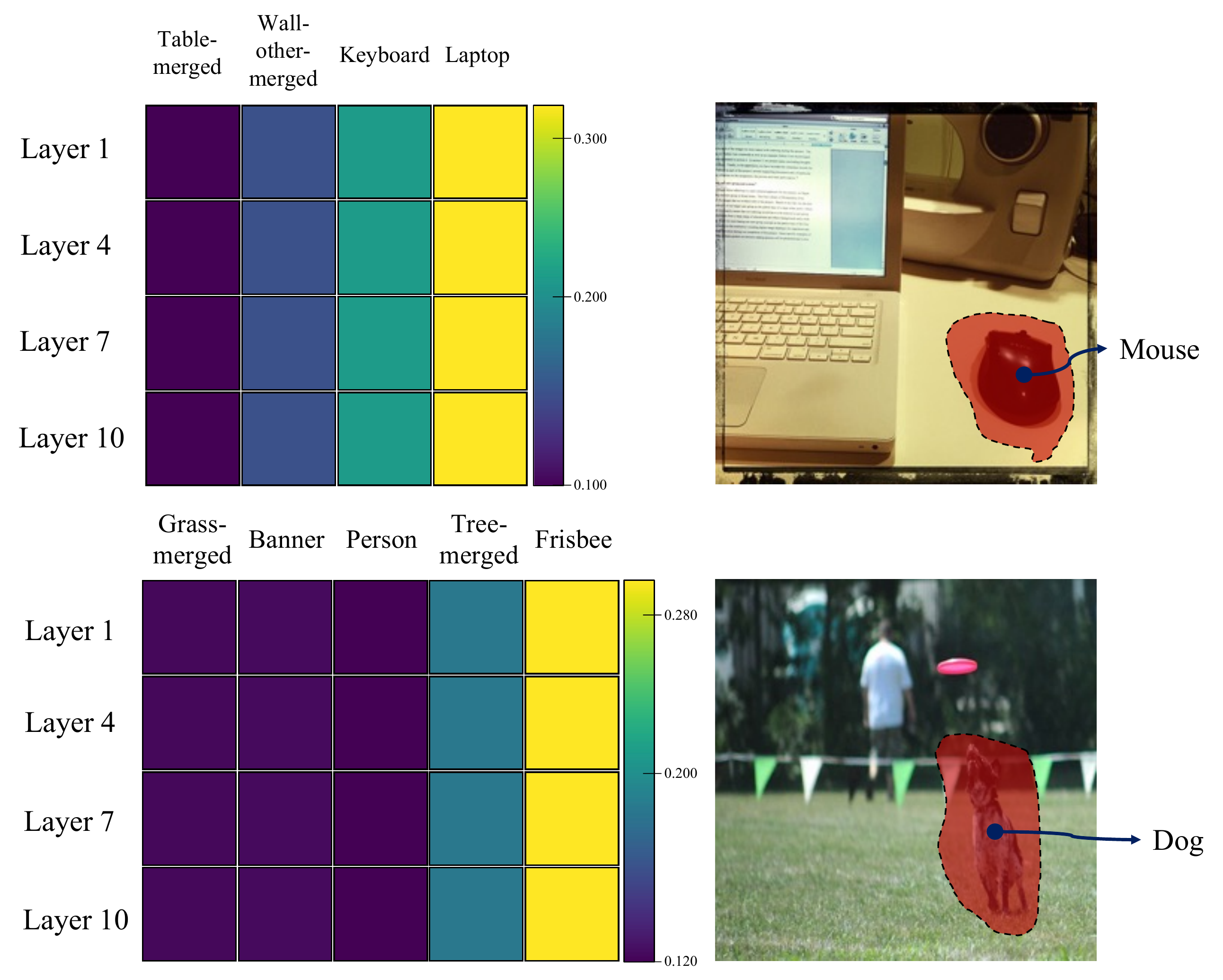}
    \caption{Self-attention visualization of the missing instance inference transformer. We visualize the self-attention of the missing instance token in the intermediate layer~(Layer 1, Layer 4, Layer 7, and Layer 10). A red transparent mask denotes the missing region of each image. Note that relevant classes of the missing region have the highest attention scores in every layer.}
    \label{fig:attn_matrix_vis}
\end{figure}

%% file: assets/figures/PositionAwarePrediction.tex
\begin{figure}[!ht]
    \centering
    \includegraphics[width=0.5\textwidth]{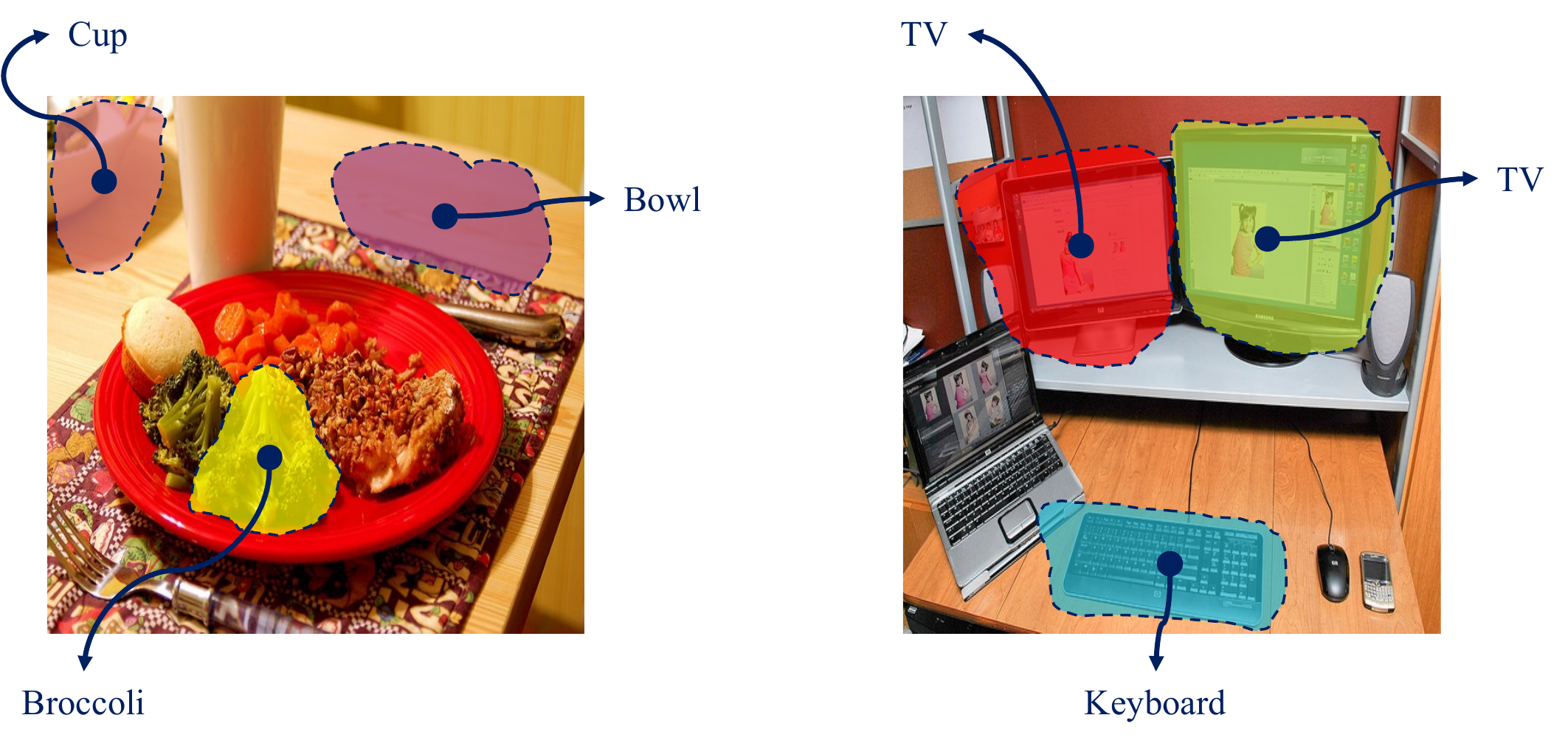}
    \caption{In this experiment, we individually hide three regions for each image and predict the class of missing instances using our approach. The predicted classes are annotated with arrows. The predicted classes can change depending on the position and shape of the missing region. }
    \label{fig:position_aware_prediction}
\end{figure}

%% file: assets/figures/MoreCOCO_1.tex
\begin{figure}[h]
    \vspace{-3.8mm}
    \includegraphics[width=0.78\textwidth]{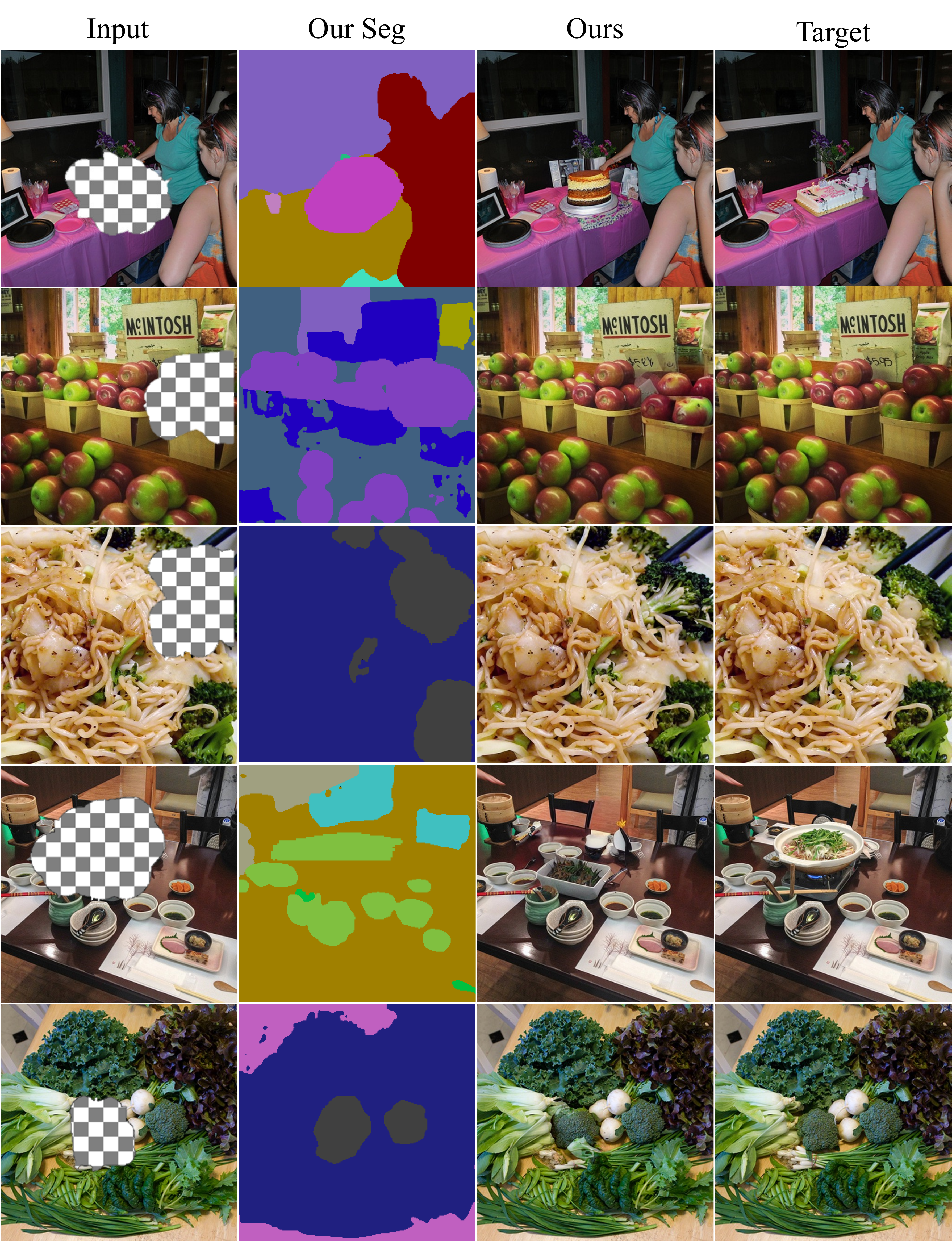}
    \centering
    \caption{Image completion results.}
    \label{fig:more_coco_1}
\end{figure}
\clearpage

%% file: assets/figures/MoreCOCO_2.tex
\begin{figure}[h]
    \includegraphics[width=0.80\textwidth]{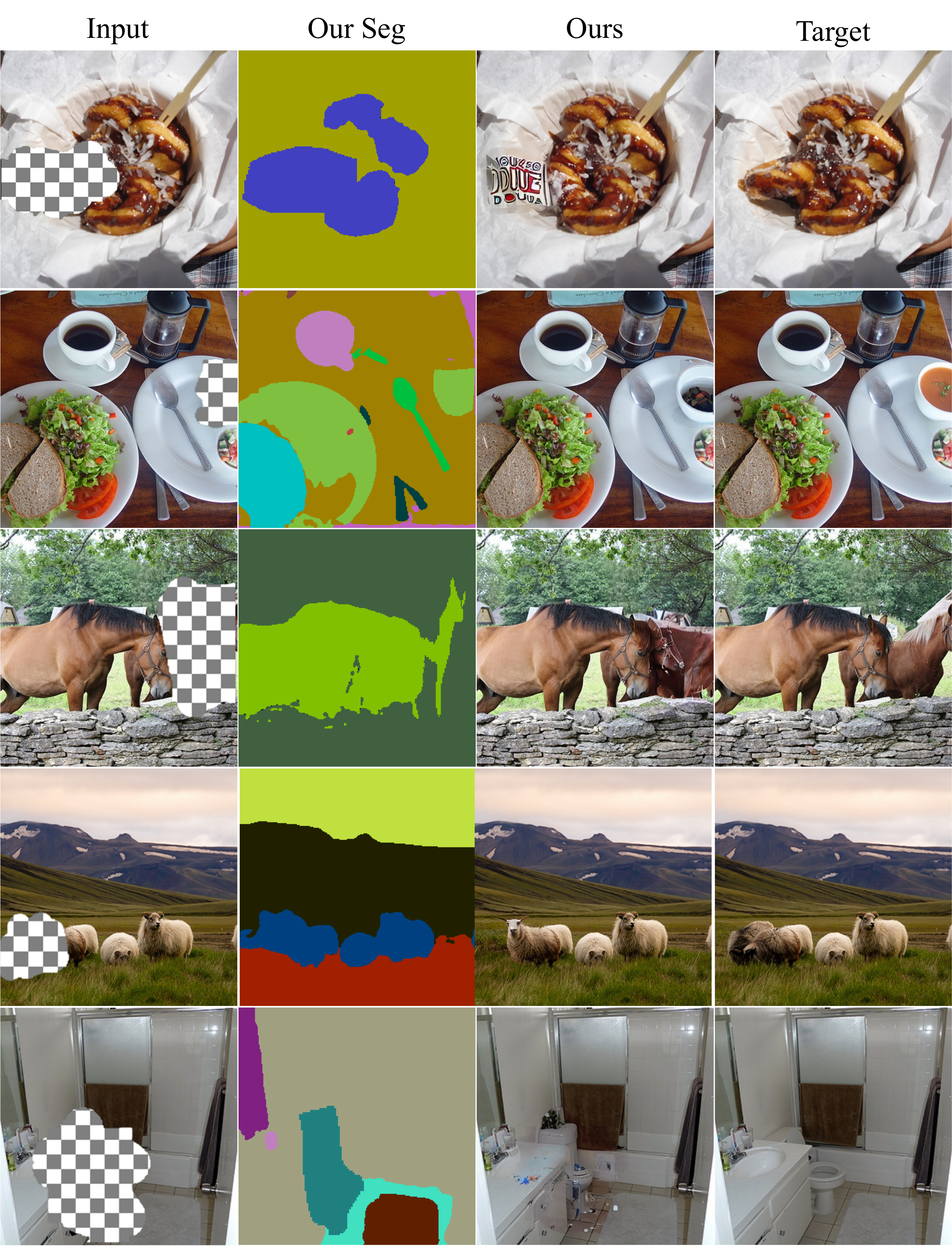}
    \centering
    \caption{Image completion results.}
    \label{fig:more_coco_2}
\end{figure}
\clearpage

%% file: assets/tables/MITArchitecture.tex
\begin{table}[!ht]
  \caption{Architecture of missing instance inference transformer.}
  \label{network:mit}
  \centering
  \resizebox{0.85\textwidth}{!}{
  \begin{tabular}{llrc}
    \toprule
    \textbf{Layer} & \textbf{Input} & \textbf{Output} & \textbf{Operation}\\
    \midrule
    Input Layer & (m, k+1, 1)  & (m, k+1, 256) & \textsc{Embedding(1,256)}\\
    \midrule
    Hidden Layer & (m, k+1, 256)  & (m, k+1, 256) & \textsc{TransEncLayer(256, 2048, 8)}\\
    Hidden Layer & (m, k+1, 256)  & (m, k+1, 256) & \textsc{TransEncLayer(256, 2048, 8)}\\
    Hidden Layer & (m, k+1, 256)  & (m, k+1, 256) & \textsc{TransEncLayer(256, 2048, 8)}\\
    Hidden Layer & (m, k+1, 256)  & (m, k+1, 256) & \textsc{TransEncLayer(256, 2048, 8)}\\
    Hidden Layer & (m, k+1, 256)  & (m, k+1, 256) & \textsc{TransEncLayer(256, 2048, 8)}\\
    Hidden Layer & (m, k+1, 256)  & (m, k+1, 256) & \textsc{TransEncLayer(256, 2048, 8)}\\
    Hidden Layer & (m, k+1, 256)  & (m, k+1, 256) & \textsc{TransEncLayer(256, 2048, 8)}\\
    Hidden Layer & (m, k+1, 256)  & (m, k+1, 256) & \textsc{TransEncLayer(256, 2048, 8)}\\
    Hidden Layer & (m, k+1, 256)  & (m, k+1, 256) & \textsc{TransEncLayer(256, 2048, 8)}\\
    Hidden Layer & (m, k+1, 256)  & (m, k+1, 256) & \textsc{TransEncLayer(256, 2048, 8)}\\
    Hidden Layer & (m, k+1, 256)  & (m, k+1, 256) & \textsc{TransEncLayer(256, 2048, 8)}\\
    Hidden Layer & (m, k+1, 256)  & (m, k+1, 256) & \textsc{TransEncLayer(256, 2048, 8)}\\
    Hidden Layer & (m, k+1, 256)  & (m, k+1, 256) & \textsc{TransEncLayer(256, 2048, 8)}\\
    \midrule
    Output Layer & (m, k+1, 256)  & (m, k+1, 256) & \textsc{FC}(256, 256), \textsc{GELU}, \textsc{LN}(256)\\
    \bottomrule
  \end{tabular}
  }
\end{table}

%% file: assets/tables/ISGArchitecture.tex
\begin{table}[h]
  \caption{Architecture for instance segmentation generator.}
  \label{network:isg_gen}
  \centering
  \resizebox{0.85\textwidth}{!}{
      \begin{tabular}{llrc}
        \toprule
        \textbf{Layer} & \textbf{Input} & \textbf{Output} & \textbf{Operation}\\
        \midrule
        Input Layer & (m,20)  & (m,20480) & \textsc{FC(20, 20480)}\\
        \midrule
        Reshape Layer & (m,20480)&(m,4,4,1280) & \textsc{Reshape}\\
        Hidden Layer & (m,4, 4, 1280)&(m,8, 8, 640) & \textsc{BigGBlock}(1280, 640, True, 20, 128) \\
        Hidden Layer & (m,8, 8, 640)&(m,16, 16, 320) & \textsc{BigGBlock}(640, 320, True, 20, 128) \\
        Hidden Layer & (m,16, 16, 320)&(m,32, 32, 160) & \textsc{BigGBlock}(320, 160, True, 20, 128) \\
        Hidden Layer & (m,32, 32, 160)&(m,32, 32, 160) & \textsc{Self-Attention} \\
        Hidden Layer & (m,32, 32, 160)&(m,64, 64, 80) & \textsc{BigGBlock}(160, 80, True, 20, 128) \\
        Hidden Layer & (m,64, 64, 80)&(m,64, 64, 1) & \textsc{BN}, \textsc{ReLU}, \textsc{Conv(80,3, 3, 1)} \\
        \midrule
        Output Layer & (m,64, 64, 1)&(m,64, 64, 1) & \textsc{Tanh} \\
        \bottomrule
      \end{tabular}
  }
\end{table}

\begin{table}[h]
  \caption{Architecture for instance segmentation discriminator.}
  \label{network:isg_dis}
  \centering
  \resizebox{0.85\textwidth}{!}{
      \begin{tabular}{llrc}
        \toprule
        \textbf{Layer} & \textbf{Input} & \textbf{Output} & \textbf{Operation}\\
        \midrule
        Input Layer & (m, 64, 64, 1)  & (m, 32, 32, 80) & \textsc{DBlock}(3, 80, True)\\
        \midrule
        Hidden Layer & (m, 32, 32, 80)  & (m, 32, 32, 80) & \textsc{Self-Attention}\\
        Hidden Layer & (m, 32, 32, 80)  & (m, 16, 16, 160) & \textsc{DBlock}(80, 160, True)\\
        Hidden Layer & (m, 16, 16, 160)  & (m, 8, 8, 320) & \textsc{DBlock}(160, 320, True)\\
        Hidden Layer & (m, 8, 8, 320)  & (m, 4, 4, 640) & \textsc{DBlock}(320, 640, True)\\
        Hidden Layer & (m, 4, 4, 640)  & (m, 4, 4, 1280) & \textsc{DBlock}(640, 1280, False)\\
        Hidden Layer & (m, 4, 4, 1280)  & (m, 1280) & \textsc{ReLU}, \textsc{GSP}\\
        \midrule
        Output Layer & (m, 1280)  & (m, 1) & \textsc{FC(1280, 1)}\\
        \bottomrule
      \end{tabular}
    }
\end{table}

%% file: assets/tables/BSCArchitecture.tex
\begin{table}[h]
  \caption{Architecture for transformer-body background segmentation completion network.}
  \label{network:bsc}
  \centering
  \resizebox{0.85\textwidth}{!}{
      \begin{tabular}{llrc}
        \toprule
        \textbf{Layer} & \textbf{Input} & \textbf{Output} & \textbf{Operation}\\
        \midrule
        Input Layer & (m, 256, 256, 1) & (m, 256, 256, 64) & \textsc{Embedding} \\
        \midrule
        Hidden Layer & (m, 256, 256, 64)  & (m, 128, 128, 128) & \textsc{Conv}(4, 2, 1)\\
        Hidden Layer & (m, 128, 128, 128)  & (m, 64, 64, 256) & \textsc{Conv}(4, 2, 1)\\
        Hidden Layer & (m, 64, 64, 256)  & (m, 32, 32, 512) & \textsc{Conv}(4, 2, 1)\\
        Hidden Layer & (m, 32, 32, 512)  & (m, 32, 32, 512) & \textsc{TransEncLayer}(512, 2048, 8)\\
        Hidden Layer & (m, 32, 32, 512)  & (m, 32, 32, 512) & \textsc{TransEncLayer}(512, 2048, 8)\\
        Hidden Layer & (m, 32, 32, 512)  & (m, 32, 32, 512) & \textsc{TransEncLayer}(512, 2048, 8)\\
        Hidden Layer & (m, 32, 32, 512)  & (m, 32, 32, 512) & \textsc{TransEncLayer}(512, 2048, 8)\\
        Hidden Layer & (m, 32, 32, 512)  & (m, 32, 32, 512) & \textsc{TransEncLayer}(512, 2048, 8)\\
        Hidden Layer & (m, 32, 32, 512)  & (m, 32, 32, 512) & \textsc{TransEncLayer}(512, 2048, 8)\\
        Hidden Layer & (m, 32, 32, 512)  & (m, 32, 32, 512) & \textsc{TransEncLayer}(512, 2048, 8)\\
        Hidden Layer & (m, 32, 32, 512)  & (m, 32, 32, 512) & \textsc{TransEncLayer}(512, 2048, 8)\\
        Hidden Layer & (m, 32, 32, 512)  & (m, 64, 64, 256) & \textsc{up}(2), ResBlock(3, 1, 1)\\
        Hidden Layer & (m, 64, 64, 256)  & (m, 128, 128, 128) & \textsc{up}(2), ResBlock(3, 1, 1)\\
        Hidden Layer & (m, 128, 128, 128)  & (m, 256, 256, 64) & \textsc{up}(2), ResBlock(3, 1, 1)\\
    
        \midrule
        Output Layer &  (m, 256, 256, 64)  & (m, 256, 256, 64) & \textsc{Reshape}, \textsc{FC(64, 64)}, GELU, LN(64)\\
        \bottomrule
      \end{tabular}
    }
\end{table}

%% file: assets/tables/SISDArchitecture.tex
\begin{table}[ht]
  \caption{Architecture for segmentation guided image completion discriminator~(OASIS version).}
  \label{network:s2i_dis}
  \centering
  \resizebox{0.85\linewidth}{!}{
  \begin{tabular}{lllll}
    \toprule
    \textbf{Operation} & \textbf{Input} & \textbf{Size} & \textbf{Output} & \textbf{Size}\\
    \midrule
    \textsc{ResBlock-Down} & $I$  &(m, 256, 256, 4)& $down_{1}$ & (m, 256, 256, 32)\\
    \textsc{ResBlock-Down} & $down_{1}$  &(m, 128, 128, 128)& $down_{2}$ & (m, 64, 64, 128)\\
    \textsc{ResBlock-Down} & $down_{2}$  &(m, 64, 64, 128)& $down_{3}$ & (m, 64, 64, 128)\\
    \textsc{ResBlock-Down} & $down_{3}$  &(m, 32, 32, 256)& $down_{4}$ & (m, 32, 32, 256)\\
    \textsc{ResBlock-Down} & $down_{4}$  &(m, 16, 16, 256)& $down_{5}$ & (m, 16, 16, 512)\\
    \textsc{ResBlock-Down} & $down_{5}$  & (m, 8, 8, 512)& $down_{6}$ & (m, 4, 4, 512)\\
    \textsc{ResBlock-Up} & $down_{6}$  & (m, 4, 4, 512)& $up_{1}$ & (m, 8, 8, 512)\\
    \textsc{Concatenate} & $down_{5}$  & (m, 8, 8, 512)& $up_{1cat}$ & (m, 8, 8, 1024)\\
                         &  $up_{1}$   & (m, 8, 8, 512)&     & \\

    \textsc{ResBlock-Up} & $up_{1cat}$  & (m, 8, 8, 1024)& $up_{2}$ & (m, 16, 16, 256)\\
    \textsc{Concatenate} & $down_{4}$   & (m, 16, 16, 256)& $up_{2cat}$ & (m, 16, 16, 512)\\
                         &    $up_{2}$  & (m, 16, 16, 256)&        &  \\

    \textsc{ResBlock-Up} & $up_{2cat}$  & (m, 16, 16, 512)& $up_{3}$ & (m, 32, 32, 256)\\
    \textsc{Concatenate} & $down_{3}$   & (m, 32, 32, 256)& $up_{3cat}$ & (m, 32, 32, 512)\\
                         &      $up_{3}$& (m, 32, 32, 256) & & \\

    \textsc{ResBlock-Up} & $up_{3cat}$  & (m, 32, 32, 512)& $up_{4}$ & (m, 64, 64, 128)\\
    \textsc{Concatenate} & $down_{2}$  & (m, 64, 64, 128)& $up_{4cat}$ & (m, 64, 64, 256)\\
                         &   $up_{4}$    &(m, 64, 64, 128)& &\\

    \textsc{ResBlock-Up} & $up_{4cat}$   & (m, 64, 64, 256)& $up_{5}$ & (m, 128, 128, 128)\\
    \textsc{Concatenate} & $down_{1}$   & (m, 128, 128, 128)& $up_{5cat}$ & (m, 128, 128, 256)\\
                         &   $up_{5}$    &  (m, 128, 128, 128)&    &\\
                         
    \textsc{ResBlock-Up} & $up_{5cat}$  & (m, 128, 128, 256)& $up_{6}$ & (m, 256, 256, 64)\\
    \textsc{Conv(3,1,1)} & $up_{6}$  & (m, 256, 256, 64)& ${I_F}$ & (m, 256, 256, N+1)\\
    \bottomrule
 \end{tabular}
  }
\end{table}

%% file: assets/tables/SISGArchitecture.tex
\begin{table}[ht]
  \caption{Architecture for segmentation-guided image completion generator~(OASIS version).}
  \label{network:s2i_gen}
  \centering
  \resizebox{1.0\linewidth}{!}{
  \begin{tabular}{lllll}
    \toprule
    \textbf{Operation} & \textbf{Input} & \textbf{Size} & \textbf{Output} & \textbf{Size}\\
    \midrule
    \textsc{Concatenate} & $I_M$ & (m, 256, 256, 3)& $I_{M_{cat}}$ & (m, 256, 256, 4)\\
                  & $M$ & (m, 256, 256, 1) & &\\
                  
    \textsc{Conv(4,1,1)} & $I_{M_{cat}}$  &(m, 256, 256, 4)& $inter_{0}$ & (m, 256, 256, 32)\\
    \textsc{Conv(3,2,1)} & $inter_{0}$  &(m, 256, 256, 32)& $inter_{1}$ & (m, 128, 128, 64)\\
    \textsc{Conv(3,2,1)} & $inter_{1}$  &(m, 128, 128, 64)& $inter_{2}$ & (m, 64, 64, 128)\\
    \textsc{Conv(3,2,1)} & $inter_{2}$  &(m, 64, 64, 128)& $inter_{3}$ & (m, 32, 32, 256)\\
    \textsc{Conv(3,2,1)} & $inter_{3}$  &(m, 32, 32, 256)& $inter_{4}$ & (m, 16, 16, 512)\\
    \textsc{Conv(3,2,1)} & $inter_{4}$  &(m, 16, 16, 512)& $inter_{5}$ & (m, 8, 8, 512)\\
    \textsc{Conv(3,2,1)} & $inter_{5}$  & (m, 8, 8, 512)& $inter_{6}$ & (m, 4, 4, 512)\\

    \textsc{Reshape, FC(4*4*512,256)} & $inter_{6}$  &(m, 4, 4, 512)& $mu$ & (m, 256)\\ 
    \textsc{Reshape, FC(4*4*512,256)} & $inter_{6}$  &(m, 4, 4, 512)& $sigma$ & (m, 256)\\
    
    \textsc{Noise Sampling, FC, Reshape} & $mu$  &(m, 256)& $z$ & (m, 8, 8, 1024)\\ 
                            & $sigma$  &(m, 256)& & \\ 
                            
    \textsc{Concatenate} & $z_{3D}$&  (m, 64, 256, 256) & $z_{y}$&(m, 64+N+1, 256, 256)\\
                            & $y$ &  (m, N, 256, 256) & &\\
                            & $M$ &  (m, 1, 256, 256) & &\\
                            
    \textsc{SPADEResnetBlock, Conv(3, 1, 1)} & $z$ &(m, 8, 8, 1024) & $up_{0}$ & (m, 8, 8, 512)\\
                              & $z_{y}$ &  (m, 64+N+1, 256, 256) & &\\

    \textsc{Concatenate} & $up_{0}$&  (m, 8, 8, 512) &  $up_{0cat}$&(m, 8, 8, 1024)\\
                            & $inter_{5}$ &  (m, 8, 8, 512) & & \\

    \textsc{UP(2), SPADEResnetBlock} & $up_{0cat}$ &(m, 8, 8, 1024) & $up_{1}$ & (m, 16, 16, 512)\\
                              & $z_{y}$ &  (m, 64+N+1, 256, 256) & &\\

    \textsc{Concatenate} & $up_{1}$&  (m, 16, 16, 512) &  $up_{1cat}$&(m, 16, 16, 1024)\\
                            & $inter_{4}$ &  (m, 16, 16, 512) & & \\

    \textsc{UP(2), SPADEResnetBlock, Conv(3, 1, 1)} & $up_{1cat}$ &(m, 16, 16, 1024) & $up_{2}$ & (m, 32, 32, 256)\\
                              & $z_{y}$ &  (m, 64+N+1, 256, 256) & &\\

    \textsc{Concatenate} & $up_{2}$&  (m, 32, 32, 256) &  $up_{2cat}$&(m, 32, 32, 512)\\
                            & $inter_{3}$ &  (m, 32, 32, 256) & & \\

    \textsc{UP(2), SPADEResnetBlock, Conv(3, 1, 1)} & $up_{2cat}$ &(m, 32, 32, 512) & $up_{3}$ & (m, 64, 64, 128)\\
                              & $z_{y}$ &  (m, 64+N+1, 256, 256) & &\\

    \textsc{Concatenate} & $up_{3}$&  (m, 64, 64, 128) &  $up_{3cat}$&(m, 64, 64, 256)\\
                            & $inter_{2}$ &  (m, 64, 64, 128) & & \\

    \textsc{UP(2), SPADEResnetBlock, Conv(3, 1, 1)} & $up_{3cat}$ &(m, 64, 64, 256) & $up_{4}$ & (m, 128, 128, 64)\\
                              & $z_{y}$ &  (m, 64+N+1, 256, 256) & &\\

    \textsc{Concatenate} & $up_{4}$&  (m, 128, 128, 64) &  $up_{4cat}$&(m, 128, 128, 128)\\
                            & $inter_{1}$ &  (m, 128, 128, 64) & & \\

    \textsc{UP(2), SPADEResnetBlock, Conv(3, 1, 1)} & $up_{4cat}$ &(m, 128, 128, 128) & $up_{5}$ & (m, 256, 256, 32)\\
                              & $z_{y}$ &  (m, 64+N+1, 256, 256) & &\\

    \textsc{Concatenate} & $up_{5}$&  (m, 256, 256, 32) &  $up_{5cat}$&(m, 256, 256, 64)\\
                            & $inter_{0}$ &  (m, 256, 256, 32) & & \\
                            
    \textsc{Conv(3,1,1), tanh} & $up_{5cat}$  & (m, 256, 256, 64)& ${I_F}$ & (m, 256, 256, 3)\\
    \bottomrule

  \end{tabular}
  }
\end{table}